\documentclass{article}

\usepackage{arxiv}

\usepackage[utf8]{inputenc} 
\usepackage[T1]{fontenc}    
\usepackage{hyperref}       
\usepackage{url}            
\usepackage{booktabs}       
\usepackage{amsfonts}       
\usepackage{nicefrac}       
\usepackage{microtype}      
\usepackage{lipsum}		
\usepackage{graphicx}
\usepackage[numbers]{natbib}

\usepackage{doi}

\usepackage{cite}
\usepackage{amsmath,amssymb,amsfonts}
\usepackage{algorithm,algorithmic}
\usepackage{graphicx}
\usepackage{textcomp}
\usepackage{tikz}
\usepackage{pgfplots}
\usepackage{tikz}
\pgfplotsset{compat=newest}
\usetikzlibrary{patterns,spy}
\usepgfplotslibrary{groupplots}
\usepackage{subcaption}
\usepackage{graphicx}
\usepackage{booktabs}
\usepackage{amsmath,amssymb}
\newcommand\rot[1]{\rlap{\rotatebox{65}{#1}}}
\usepackage{placeins}
\usepackage{multirow}
\usepackage{url}

\usepackage{array}
\usepackage{graphicx} 

\newcommand{\fm}[1]{{\color{black}{#1}}}

\title{Less is More: The Influence of Pruning on the
Explainability of CNNs}

\author{ \href{https://orcid.org/0000-0000-0000-0000}{\includegraphics[scale=0.06]{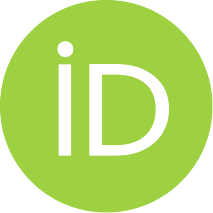}\hspace{1mm}Florian Merkle}\\
	Innsbruck University\\
	Department of Information Systems, Production \\ and Logistics Management\\
	Austria \\
	\texttt{florian.merkle@student.uibk.ac.at} \\
	\And
	David Weber \\
	MCI – The Entrepreneurial School\\
	Innsbruck\\
	Austria \\
	\texttt{weber\_david@outlook.com} \\
	\AND
	\href{https://orcid.org/0000-0000-0000-0000}{\includegraphics[scale=0.06]{orcid.pdf}\hspace{1mm}Pascal Schöttle}\\
	Josef Ressel Centre for Security Analysis of IoT Devices\\
	Innsbruck\\
	Austria \\
	\texttt{pascal.schoettle@mci.edu} \\
	\And
	\href{https://orcid.org/0000-0000-0000-0000}{\includegraphics[scale=0.06]{orcid.pdf}\hspace{1mm}Stephan Schlögl}\\
	MCI – The Entrepreneurial School\\
	Innsbruck\\
	Austria \\
	\texttt{stephan.schloegl@mci.edu} \\
	\And
	\href{https://orcid.org/0000-0000-0000-0000}{\includegraphics[scale=0.06]{orcid.pdf}\hspace{1mm}Martin Nocker}\\
	Josef Ressel Centre for Security Analysis of IoT Devices\\
	Innsbruck\\
	Austria \\
	\texttt{martin.nocker@mci.edu} \\
}

\renewcommand{\headeright}{}
\renewcommand{\undertitle}{}
\renewcommand{\shorttitle}{NN Pruning and Explainability}

\begin{document}
\maketitle

\begin{abstract}
	Over the last century, deep learning models have become the state-of-the-art for solving complex computer vision problems. These modern computer vision models have millions of parameters, which presents two major challenges: (1) the increased computational requirements hamper the deployment in resource-constrained environments, such as mobile or IoT devices, and (2) explaining the complex decisions of such networks to humans is challenging. Network pruning is a technical approach to reduce the complexity of models, where less important parameters are removed. The work presented in this paper investigates whether this reduction in technical complexity also helps with perceived explainability. To do so, we conducted a pre-study and two human-grounded experiments, assessing the effects of different pruning ratios on explainability. Overall, we evaluate four different compression rates (i.e., 2, 4, 8, and 32) with 37\,500 tasks on Mechanical Turk. Results indicate that lower compression rates have a positive influence on explainability, while higher compression rates show negative effects. Furthermore, we were able to identify sweet spots that increase both the perceived explainability and the model's performance.
\end{abstract}


\section{Introduction}
\label{sec:introduction}

Today's products and services increasingly benefit from integrating ever more powerful machine learning (ML) features. In particular, the use of deep learning and respective deep neural networks (DNN) has significantly expanded upon the capabilities of intelligent systems and consequently improved the performance of autonomous vehicles, virtual assistants, fraud detection software, and tools that make heavy use of image recognition technology.
Concrete application domains for these DNNs are found, e.g., in  
medicine~\citep{ker2017deep}, production~\citep{weber2021}, cyber security~\citep{alazab2019deep} and finance~\citep{huang2020deep}. To this end, convolutional neural networks (CNN) -- a particular type of DNN -- have demonstrated great performance in computer vision tasks. This includes image classification of chest x-rays to treat COVID-19 patients~\citep{dey2021covid}, image segmentation of MRI scans to analyze different regions of the brain~\citep{ali2019application}, or object detection to support autonomous driving~\citep{pathak2018application}.

\begin{figure*}
    \centering
  \includegraphics[width=.8\textwidth]{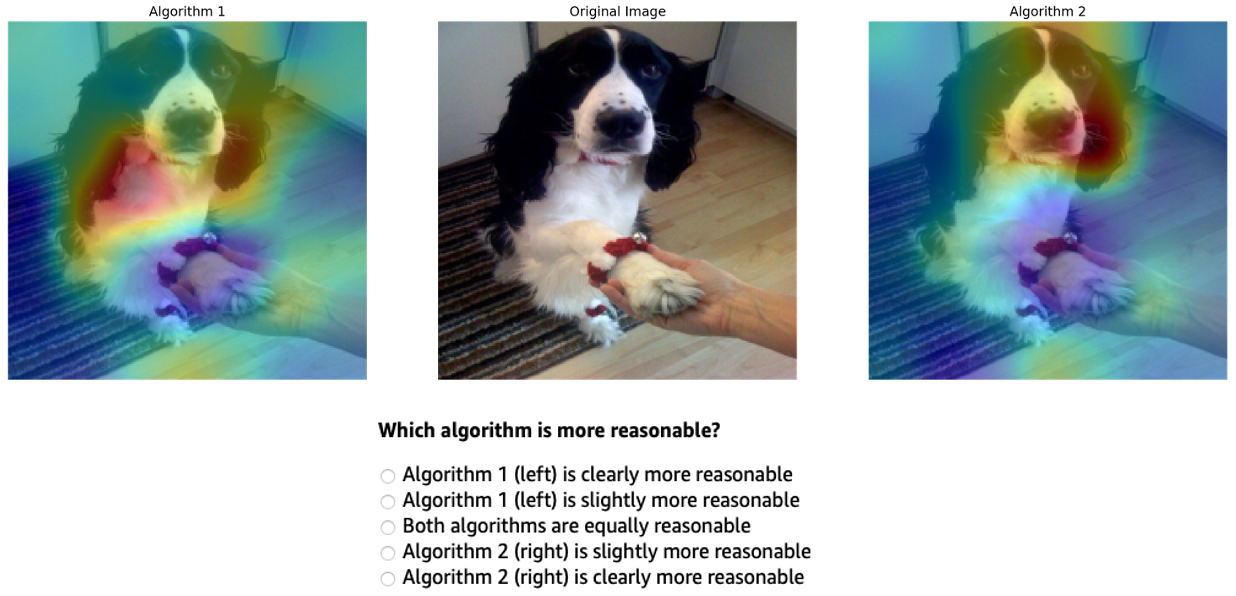}
  \caption{Which algorithm is more reasonable? In the middle, we show the original picture; on the left, the explainability heat map of compression rate 1; and on the right, the explainability heat map for compression rate 8. Red colors indicate more important regions, while blue colors indicate less important regions.}
  \label{fig:exsetup}
\end{figure*}

\fm{
These recent advancements in computer vision can be largely attributed to the increased availability of data and significant improvements in computational power, enabling the development of larger and more complex models~\citep{han2021pre}. This remains less problematic when energy and computational resources are readily available, i.e., the computation is performed on a workstation or in the cloud. However, computer vision systems are also deployed in resource-constrained settings such as mobile and smart devices or Internet of Things (IoT) applications. Moreover, many of these applications, such as smart doorbells and cameras~\citep{lulla2021iot}, are privacy- or security-sensitive and rely on a certain level of trust from the users. Prior research indicates that explainable decisions increase users' trust towards deep learning model~\citep{ribeiro2016should}.

Therefore, this study aims to tackle the following limitations that today's DNNs (and consequently CNNs) face:}

\begin{enumerate}
    \item Complexity: State-of-the-art DNNs have to deal with millions of parameters and thus require large amounts of computing power and memory. This is especially important for applications in resource-constrained environments, such as smartphones and IoT devices. Furthermore, more parameters negatively influence the inference time, which is critical for real-time applications such as autonomous cars or face detection. One technical approach to retrospectively reduce DNN parameters is so-called neural network (NN) pruning, where less important parameters are deleted.
    \item Explainability: DNNs experience a lack of explainability that leaves little understanding of why a particular decision was made~\citep{du2019techniques}. Especially the structure of CNNs, consisting of complex internal relations, can be challenging to explain~\citep{arrieta2020explainable}. Understanding the reasoning of these systems is crucial, especially for high-stake decision-making and highly regulated domains. A DNN's decisions may determine the difference between life and death, as in healthcare, medicine, or autonomous driving. More explainable, reasonable, and transparent DNNs would increase trust, acceptance, and awareness in society.
\end{enumerate}

Intuitively, more parameters, i.e., more complexity, lead to a lower explainability, as not every connection can be interpreted with human reasonability. 
This is clearly shown with DNNs, as their high number of parameters and complex internal sequences appear opaque to humans~\citep{burkart2021survey} and certain explainability methods are prone to produce noisy and indistinct explanations~\citep{khakzar2019improving}.
At the same time, this striving for complexity helped state-of-the-art NNs reach their current performance. In this paper, we start from the hypothesis that retrospectively reducing the number of parameters with the help of NN pruning can increase the explainability to humans.

\begin{figure}[th]
    \centering
    \resizebox{.5\linewidth}{!}{

\definecolor{color0}{RGB}{0, 73, 131} 
\definecolor{color4}{RGB}{244, 155, 0} 
\colorlet{color1}{color0!75}
\colorlet{color2}{color4!25}
\colorlet{color3}{color4!75}

\begin{tikzpicture}
\pgfplotsset{
    width=.95\linewidth,
	height=.5\linewidth,
	tick align=outside,
	scaled y ticks=false,
	major tick length=1.5mm,
	xmin=0.75, xmax=6.25, 
}
\begin{axis}[
    title={},
    axis y line*=left,
    xlabel={Compression Rate (CR)},
    ylabel={Accuracy},
    ymin=0.8, ymax=1.0,
    xtick={1,2,3,4,5,6},
    xticklabels = {1,2,4,8,16,32},
    ytick={0.8,0.825,0.85,0.875,0.9,0.925,0.95,0.975,1.0},
    yticklabels = {80.0,82.5,85.0,87.5,90.0,92.5,95.0,97.5,100},
    xmajorgrids=true,
    ymajorgrids=true,
    grid style=dashed,
    xtick pos=left,
    ytick pos=left,
]


\addplot[
    color=color0,
    mark=*,
    thick
    ]
    coordinates {
    (1,0.97)
    (2,0.972857)
    (3,0.959285)
    (4,0.9410713)
    (5,0.9339285)
    (6,0.9007142)
    };\label{plot_one}

\addplot[
    color=color1,
    mark=square,
    thick,
    ]
    coordinates {
    (1,0.8585)
    (2,0.8640)
    (3,0.8208)
    (4,0.8120)
    (6,0.8132)
    };\label{plot_two}
\end{axis}
\begin{axis}[
   title={},
  axis y line*=right,
  axis x line=none,
  ymin=-.05, ymax=.05,
  ylabel=Explainability,
  ytick={-.05, -0.025,0,.025,.05},
  yticklabels = {-0.050, -0.025,0,0.025,0.500},
    legend style={at={(.0, 1.6)},anchor=north west},
]
 \addlegendimage{/pgfplots/refstyle=plot_one}\addlegendentry{Top 1-test accuracy}
 \addlegendimage{/pgfplots/refstyle=plot_two}\addlegendentry{Human rater accuracy}
\addplot[thick,mark=x,color4]
  coordinates{
    (1,-0.0168)
    (2,0.0071)
    (3,-0.0055)
    (4,0.0152)
}; \addlegendentry{Explainability index}
\end{axis}
\end{tikzpicture}}
    \caption{Top-1 test-set accuracies (dark blue, left y-axis), human rater accuracies (light blue, left y-axis), and our explainability measure (orange, right y-axis) for different compression rates.}
    \label{fig:pruningacc}
\end{figure}
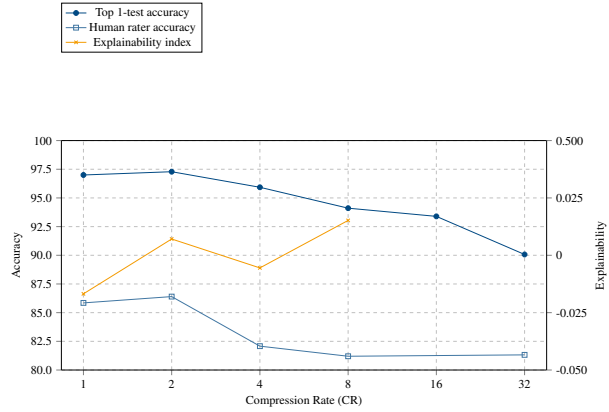

We apply several network pruning compression rates (CR), i.e., CR 2, 4, 8, and 32, to a VGG-16~\citep{simonyan2014deep} network. VGG-16 is a common CNN architecture with a simple and homogeneous structure. Its size of 138 million parameters provides ideal preconditions for pruning and opaque behavior. Grad-CAM~\citep{selvaraju2020grad} is our explainability method of choice, as it allows for a visual explanation for the internal reasoning of black-box CNNs, passes several sanity checks by \citet{adebayo2018sanity}, and applies to any CNN architecture.
We conduct three human-grounded experiments with 37\,500 tasks on Amazon's Mechanical Turk\footnote{Online: \url{https://www.mturk.com} [accessed: \today]} platform. We analyze the data gathered from these experiments with evaluation metrics from both objective and subjective experiments, i.e., one where ground truth is given and one where ground truth is not given.
Figure~\ref{fig:exsetup} shows the experimental setup of one of our experiments. Our results indicate a sweet spot of mild pruning, with a CR of 2, i.e., a CNN with half the parameters. As visible in Figure~\ref{fig:pruningacc}, not only does the top 1 test accuracy of the CNN improve, but also the evaluation metrics from both subjective (explainability index) and objective experiments (human-rater-accuracy).
The remainder of this paper reports the details of this investigation.

We discuss the relevant theoretical background in Section~\ref{sec:relatedwork}. Next, we describe our methodological approach in Section~\ref{sec:methodology}. Section~\ref{sec:results} then reports on respective results, Section~\ref{sec:discussion} reflects on the limitations of our approach, and Section~\ref{sec:limitations} highlights directions for future work. Finally, Section~\ref{sec:conclusion} summarizes and concludes the paper. 

\section{Theoretical Background}
\label{sec:relatedwork}
ML methods, particularly DNN approaches, are powerful tools for creating predictions based on data. Yet, they are often called black-box models, as their inner workings lack transparency. 
Even so, these non-transparent systems are increasingly used for high-stake decision-making in healthcare, precision medicine, criminal justice, autonomous driving, and other highly regulated domains impacting human lives~\citep{rudin2018please9,angelov2020towards,tjoa2020survey}. Not being able to explain the system's decisions thus poses evident dangers~\citep{adadi2018peeking16}. This implies that it is not enough to receive a prediction (i.e., the \textit{what}) but also the explanation of how this prediction was made (i.e., the \textit{why}). Or, as \citet{doshi2017towards} emphasize, \textit{``Explainability completes the problem formulation''}. 

\subsection{Explainability in Machine Learning}
Unfortunately, ML literature lacks a clear definition of \textit{explainability} and \textit{interpretability}. Consequently, terms are regularly ill-defined, misused or referred to in different ways~\citep{lipton2018mythos}.
Also, it may happen that \textit{explainability} and \textit{interpretability} are used synonymously~\citep{cabitza2019new,miller2019explanation}, although \citet{gilpin2018explaining} clearly state that \textit{interpretability} (i.e., human-understandability) and \textit{fidelity} (i.e., the accurate description of a system's internal workings) are required to reach \textit{explainability}.
Several definitions for \textit{interpretability} exist. Some refer to it as the ability to explain technology in human-understandable terms~\citep{doshi2017towards,molnar2020interpretable}, which in turn brings forth the question of what understandable means.
\citet{ras2018explanation29}, on the other hand, define interpretability as the extent to which \textit{``a user is able to obtain true insights into how actionable outcomes are obtained''} and split the concept into the subproperties \textit{clarity} and \textit{parsimony}~\citep[p.~6]{ras2018explanation29}. High \textit{clarity} is achieved when the explanation is unambiguous, while high \textit{parsimony} is given when the user perceives it as simple, which depends on their capabilities.
\textit{Fidelity} generally describes whether an explanation is accurate. \citet{kulesza2013too30} divide fidelity further into \textit{soundness} and \textit{completeness}, where \textit{soundness} describes whether the explanation is correct and faithful to the task model, and \textit{completeness} is achieved when the explanation covers the entire dynamic of the task model.

ML explainability supports the social acceptance of, trust in, and social interaction with ML systems and fosters their safety and knowledge acquisition~\citep{ribeiro2016should}. That is, the systems' safety may be increased through increasing its explainability~\citep{miller2019explanation}, and so is the possibility for identifying faulty behavior, as more explainable systems may ease testing, auditing, and debugging~\citep{molnar2022}. Furthermore, explainability helps people to successfully interact with ML systems and eventually reach intended goals~\citep{molnar2022}. 
In short, ML explainability supports researchers and practitioners alike and, at the same time, helps extract the `knowledge' a system uses~\citep{molnar2022}. 
Finally, the European Union's \textit{Right to Explanation}~\citep{goodman2017european} demonstrates the regulatory importance of humanly explainable ML systems, aiming to achieve equality and unbiased decision-making by algorithms~\citep{gilpin2018explaining}. \citet{cheng2021socially} support these efforts by defining the concept of \textit{Socially Responsible AI Algorithms} and providing four fundamental responsibilities: functional, legal, ethical, and philanthropic.

\subsection{Methods of CNN Explainability}
\label{sec:xAImethods}

\citet{guidotti2018survey} describe the \textit{black-box explanation problem} as the challenge to provide an explanation of the black-box model through an interpretable system or method. The difficulty lies in providing insights about the internal processes that lead to a DNN's prediction and further clarifying under which circumstances they can be trusted~\citep{ras2018explanation29} and producing insights into model predictions~\citep{ras2022explainable}. DNNs may have millions of parameters, making it hard to analyze their internal representations and the respective information flow throughout the network. Their complex learning procedure is determined by many components, including regularization, activation, and loss functions.
Specifically, CNNs entail complex internal relations due to their structure. They consist of sequences of convolutional and pooling layers that learn increasingly higher-level features. The challenge here is to reduce the complexity of these operations, which is usually done by visualizing saliency maps~\citep{gilpin2018explaining}. In theory, it should be easier for humans to understand CNNs than regular DNNs that do not make use of convolutional layers, as our cognitive skills favor the understanding of this type of visual data~\citep{arrieta2020explainable}. This assumption has led to the definition of model-agnostic~\citep{lundberg2017unified,ribeiro2016should} as well as model-specific methods, both of which aim to help with explanations for CNN decisions.
Model-agnostic methods are applicable to any ML algorithm and are usually applied after the model has been trained -- post-hoc \citep{carvalho2019machine}. While these methods analyze pairs of feature input and output, they do not have access to the inner information of the analyzed models \citep{molnar2020interpretable}. Common methods in use are Local Interpretable Model-Agnostic Explanations (LIME) \citep{ribeiro2016should} and Shapley Values (SHAP) \citep{lundberg2017unified}.
Model-specific methods either map the output back to the given input or explain the representation of the external world inside the layers~\citep{arrieta2020explainable,gilpin2018explaining}. 
Most of the model-specific methods fall into the first category. Perturbation-based methods deliver good estimates of the input pixels' impact yet induce high computational costs~\citep{ancona2017towards}. Hence, backpropagation-based methods are commonly used, e.g.,~\citep{sundararajan2017axiomatic,shrikumar2017learning,zhou2016correlation}, as they compute the attributions with one or few forward and backward passes resulting in less computational costs~\citep{ancona2017towards}. Most backpropagation-based techniques achieve a balance in visualizing areas of high network sensitivity and high network activation~\citep{gilpin2018explaining}. One of the most prominent approaches here is Gradient-Weighted Class Activation Mapping (Grad-CAM) proposed by \citet{selvaraju2017grad}. Grad-CAM uses Class Activation Mapping (CAM), originally proposed by Zhou et al.~\citep{zhou2016learning}, and visualizes the regions of an input image that are important for the model's prediction by using the class-specific gradient information. Further, this method applies to any CNN-based architecture and does not alter the architecture in any way. 
\subsection{Evaluating Explainability}
Evaluating ML explainability has a two-fold goal~\citep{markus2021role}: First, it assesses if explainability is achieved. Here, the focus of the evaluation lies in determining whether the provided explainability method achieves the defined objective~\citep{lipton2018mythos}. Second, it aims to formally compare available explainability methods and consequently identify preferences. One of the biggest challenges therefore lies in the evaluation itself, as no ground truth is given~\citep{samek2019explainable,ras2022explainable}. This is especially true for post-hoc explainability methods, where one attempts to explain the inner workings of a black-box model. Finally, the ultimate target is to assess to what extent all properties of explainability are satisfied~\citep{markus2021role}.

There are generally two factors that determine whether an ML model is understandable~\citep{zhou2021evaluating}, i.e., the human's understanding given through capacity, and the model's features. Evaluating explainability is thus a result of combining these two factors.
To this end, \citet{doshi2017towards} describe three categories of explainability evaluation approaches. 

Application-grounded evaluation measures the quality of an explanation in the context of its intended task, such as whether it results in less discrimination or better error identification. The benefit lies in testing to what extent the explainability method is helpful to the user~\citep{ras2022explainable}. Exemplary experiments include domain expert experiments with identical or simpler application tasks.
For instance, if the task is to diagnose a particular disease, the most ideal way to demonstrate the model's workings is to have doctors perform the diagnosis. A good indication of explainability is how well they explain a decision~\citep{doshi2017towards}.

The human-grounded evaluation assesses simpler human-subject experiments while maintaining the essence of the target application. This has several benefits, especially when the goal is to test general notions of explanation quality. Also, it is less expensive and may use a larger subject pool, as participants do not require domain expertise.
Possible experiments include a binary forced choice, forward simulation, and counterfactual simulation. For instance, one might ask a user to simply choose the best fitting explanation from a pool of explanations~\citep{doshi2017towards}.

Functionally-grounded evaluation does not demand human experiments but instead uses a formal definition of explainability as a proxy for explanation quality. Hence, it is most appropriate when a class of methods has already been evaluated, when a method is not yet mature enough, or when human-subject experiments would be unethical. The approach benefits from lower time and cost requirements, as no human-subject experiments are necessary~\citep{doshi2017towards}.
However, \citet{carvalho2019machine} argue that results from a functionally-grounded approach have low validity, as human feedback is missing and the defined proxies may not fully measure explainability.

\subsection{Evaluation Metrics}
With application- and human-grounded evaluations, selecting the correct evaluation metrics plays a critical role in correctly evaluating a method. To this end, \citep{zhou2021evaluating} differentiate between \textit{subjective} and \textit{objective} metrics. Subjective metrics are surveyed during, or after a task to gather the user's subjective response. They include trust, confidence, preference, or reasonability, and as such have been used in a variety of previous evaluations (e.g., \citep{ribeiro2016should,zhou2019effects,zhou2016correlation}). Objective metrics are surveyed before, during, or after a task. They include human metrics, such as physiological and behavioral indicators, informed decision-making, task time length, or task performance. For instance, \citet{schmidt2019quantifying} demonstrate that faster and more accurate decisions regularly indicate an intuitive understanding of explanations.

Functionally-grounded evaluation metrics, on the other hand, consist of various quantitative metrics to objectively assess the quality of an explanation, or more specifically whether certain explainability axioms are met~\citep{zhou2021evaluating}. Benchmarks without human intervention also fall under this category~\citep{ras2022explainable}. Examples contain model size~\citep{guidotti2018survey}, remove and retrain (ROAR)~\citep{hooker2018benchmark}, diversity~\citep{nguyen2020quantitative}, sanity checks~\citep{adebayo2018sanity}, or interaction strength~\citep{markus2021role}. 

While many quantitative metrics are proposed, a general computational benchmark across all possible explainability methods is difficult~\citep{nguyen2020quantitative}, as explainability is still a subjective concept where the perceived quality is user- and task-dependent.

\subsection{Neural Network Pruning}
\label{susec:nnpruning}
 
\begin{algorithm}[H]
\caption{Generic pruning algorithm}
\label{alg:high_level_pruning}
\begin{algorithmic}
\REQUIRE $N$: number of iterations; $x$: data set
\STATE Initialize $W$
\STATE Train $f(x; W)$ to convergence
\STATE Set $M$ to ones
\FOR{$i$ in $N$}
    \STATE Prune $M$ according to selection criterion
    \STATE Fine-tune model $f(x; W)$ to convergence
\ENDFOR
\ENSURE $M$: pruning mask; $W$: fine-tuned model weights
\end{algorithmic}    
\end{algorithm}

\fm{NN pruning describes the reduction of network parameters to decrease the computational requirements and enhance the energy efficiency in constrained environments such as mobile or IoT devices~\citep{widmann2023pruning} or for large foundational models~\citep{cheng2024survey}. }Modern NNs are typically over-parameterized for the task at hand, leading to extensive redundancies in the model~\citep{liu2018rethinking,JIANG202121}.  
The goal of pruning is to reduce these redundancies and memory requirements, which ultimately helps save computational resources. 
While the idea of network pruning was initially introduced in the late 1980s~\citep{lecun1989optimal}, it is the emergence of deep learning and the consequent rise in memory and storage requirements~\citep{blalock2020state}, which has recently led to increased interest in the concept. 
Moreover, it has been shown that a careful selection of the to-be-removed parameters does not only reduce the resource requirements of a model, but may even increase its accuracy~\citep{frankle2018lottery,han2015learning}, robustness against adversarial attacks~\citep{merkle2021pruning}, and energy efficiency~\citep{widmann2023pruning}. Research has shown that using NNs with a lower number of parameters from the start achieves lower accuracy values than using a larger model and pruning it retrospectively~\citep{han2015learning}.
Furthermore, pruning can also be combined with other methods such as knowledge distillation, to increase the amount of compression without breaking convergence~\citep{Aghli2021}.  
Pruning approaches may be described along five dimensions:

(1) The \textbf{selection criterion} defines how to select the parameters to be pruned. Many approaches have been proposed, e.g., magnitude-based (on absolute values)~\citep{han2015learning}, based on the gradients~\citep{blalock2020state}, the Hessian matrix of the loss function~\citep{lecun1989optimal}, or based on the $L_2$ norm of the network structure~\citep{he2018soft}.
Network pruning can also be incorporated into the DNN's learning procedure~\citep{huang2018data} or be formulated as its own optimization problem~\citep{zhang2018systematic}. Finally, random pruning often serves as a baseline and sanity check~\citep{blalock2020state,frankle2018lottery}. 

(2) The \textbf{scope} determines whether the selection process is performed locally~\citep{he2018soft}, where each layer is pruned separately, or globally, where all weights are considered simultaneously for the selection process. Global network pruning results in different sparsity levels in each layer, while local pruning yields a steady pruning ratio over the whole network~\citep{blalock2020state}.

(3) \textbf{Scheduling} defines when pruning is conducted. Most methods, e.g.,~\citep{han2015learning}, apply pruning after the training. The network is either pruned in one step to the desired compression rate~\citep{liu2018rethinking}, called one-shot pruning, or an iterative process of pruning and consequent training is applied~\citep{han2015learning,gale2019state}.

(4) The \textbf{pruning-structure} describes the granularity of a method,
where the unstructured approach prunes single weights~\citep{lecun1989optimal,blalock2020state}, while structured pruning removes entire parts, such as kernels and filters~\citep{he2018soft}, or even whole residual blocks~\citep{huang2018data}. Since the first approach produces sparse matrices of the same size as the unpruned network, dedicated hardware is necessary to accomplish practical improvements.

(5) \textbf{Fine tuning} refers to the training phase, which happens after pruning is applied.
Here, recent work explores whether retraining a pruned network from scratch, using a new set of randomly initialized values, would lead to a different accuracy~\citep{liu2018rethinking} than fine-tuning the remaining weights with their pre-pruning values~\citep{han2015learning}.
To this end, the lottery ticket hypothesis proposed by \citet{frankle2018lottery} suggests that a pruned network can reach a higher accuracy than its unpruned equivalent when retrained from scratch using its initial (random) values.

The number of parameters that are removed by NN pruning is determined by the compression rate, where a CR of 1 stands for the unpruned NN, while a CR of 2 yields a NN with only half of the original parameters, a CR of 4 that the resulting NN has only a fourth of the original NN's parameters and so forth. 

While other variations do exist, Algorithm~\ref{alg:high_level_pruning} 
covers most pruning methods \citep{blalock2020state}. The procedure starts with an untrained model $f(x;W_0)$ and returns a pruned model $f(x;M \odot W')$ that has been fit to the training data $x$. $W$ represents the model's weights with $M \in \{0,1\}^{|W'|}$ being a binary mask with the exact shape of $W$. By setting a value in $M_{i,k}$ to zero, its correspondent weight $W_{i,k}$ is effectively pruned as its value is always zero after the element-wise multiplication of $W$ and $M$. Equally, the weight's gradient is always zero, preventing it from taking on a non-zero value in the following learning iterations.

\subsection{Neural Network Pruning to Increase Explainability}
Previous work on leveraging NN pruning for ML explainability is scarce. \citet{khakzar2019improving} argue that current gradient-based attribution methods produce noisy results due to the complexity of current model architectures. 
Yet, if one uses input-specific pruning, where only neurons with high predictive contributions are kept, the global importance information of the attribution method may be improved. Their approach, called PruneGrad, differs from traditional model pruning in (1) that it does not take the whole dataset into account, (2) that it does not decrease the memory footprint, and (3) that it does not increase the inference speed. It is shown that roughly 50\% of neurons can be removed without any changes in the output, while excessive pruning (i.e., over 80\%) also removes highly contributing neurons and thus results in significant output changes. To perform a functionally-grounded evaluation of their PruneGrad method, \citet{khakzar2019improving} apply the sanity checks by \citet{adebayo2018sanity}, the pixel perturbation benchmark by \citet{samek2016evaluating}, and the ROAR framework by \citet{hooker2018benchmark}. Evaluation results consistently outperform other gradient-based attribution methods. 

On the other hand, \citet{abbasi2017interpreting} extend previously proposed filter importance indices to visually apply filter pruning. Their structured algorithm prunes filters with visually redundant pattern selectivity, thereby increasing the explainability of the CNN. As a result, memory savings and smaller computational costs are reached while making the CNNs more explainable.

Other scientific contributions leverage explainability methods as a means for model pruning. For example, the work of \citet{dotter2018visualizing} explores the suitability of datasets for certain models once they are pruned. Doing so, they show that visualizing a sample dataset of the final convolutional layer for different pruning ratios helps in making class separability visually understandable. Building on this, \citet{zhang2021channel} were the first to use explainability theory to guide channel pruning. 

Other works apply different model explainability methods as CNN pruning criteria, as it is a challenge 
to identify criteria by which the importance of parameters can be measured. \citet{yeom2021pruning} and \citet{soroush2023compressing} use Layer-wise relevance propagation (LRP)~\citep{bach2015pixel}. Their results show that the novel LRP criterion is not only comparable to state-of-the-art but outperforms previous criteria in transfer-learning scenarios. \citet{sabih2020utilizing} utilize DeepLIFT~\citep{shrikumar2017learning} to obtain the importance of certain neurons for NN pruning and the quantification of NN weights. With this, they aim to address a broad range of pruning methods, including structured, unstructured, CNN filter, and neuron pruning. 
\citet{yao2021deep} propose an explainability-based filter pruning framework based on activation maximization~\citep{zeiler2014visualizing}. By visualizing every filter with activation maximization, they find that over 50\% of the filters contain either repetitive or no information, making them redundant or invalid. The filters are then pruned based on a filter similarity matrix, which measures color and texture similarities. 
\fm{Related to the exploitation of explainability approaches for pruning, \citet{cheng2024survey} call for the development of explainable pruning methods in their survey paper. First efforts in this direction include the work of \citet{yu2023x} for vision transformer architectures or the method of \citet{rajapaksha2024explainable} for large language models.}
Frankle and Bau~\citep{frankle2019dissecting} examine the interpretability, quantitatively measured with the network dissection technique~\citep{bau2017network}, of the ResNet50 pruned with the lottery ticket procedure~\citep{frankle2018lottery} and find that network pruning has no influence for moderate compression rates. Finally, Arazo et al.~\citep{arazo2024xpression} propose an evaluation metric that combines a model's explainability, measured by the computable infidelity metric~\citep{yeh2019fidelity} with its compression rate in order to optimize the two objectives simultaneously.

Summarizing, we may argue that NN explainability methods are far more utilized to guide NN pruning than NN pruning is utilized to support NN explainability, although previous work has shown that there is merit in the latter. Thus, the goal of our work is to focus on NN pruning for ML explainability and measure its suitability using human-grounded evaluation.

\section{Methodology}
\label{sec:methodology}

The goal of our work is to investigate the influence of NN pruning on the explainability of CNNs. In particular, we focus on the compression rates (CRs, cf. Sec.~\ref{susec:nnpruning}) as a variable to indicate the extent of NN pruning and assess its effects on perceived CNN explainability. In this sense, our methodology may be described as a three-phase experimental study, i.e., pre-study, Experiment 1, and Experiment 2. 
Following, we first describe the technical setup we used and then provide more details on the three human-grounded experiments.

\subsection{Technical Setup}
\label{sec:technical_setup}
This section presents the technical details of the chosen model, dataset, pruning approach, and explainability method, along with the rationale for these selections.

\subsubsection{Dataset and CNN Architecture}
\label{sususec:architecture}

\begin{figure}[tbh]
    \centering
    \resizebox{\linewidth}{!}{
    \includegraphics[scale=0.35]{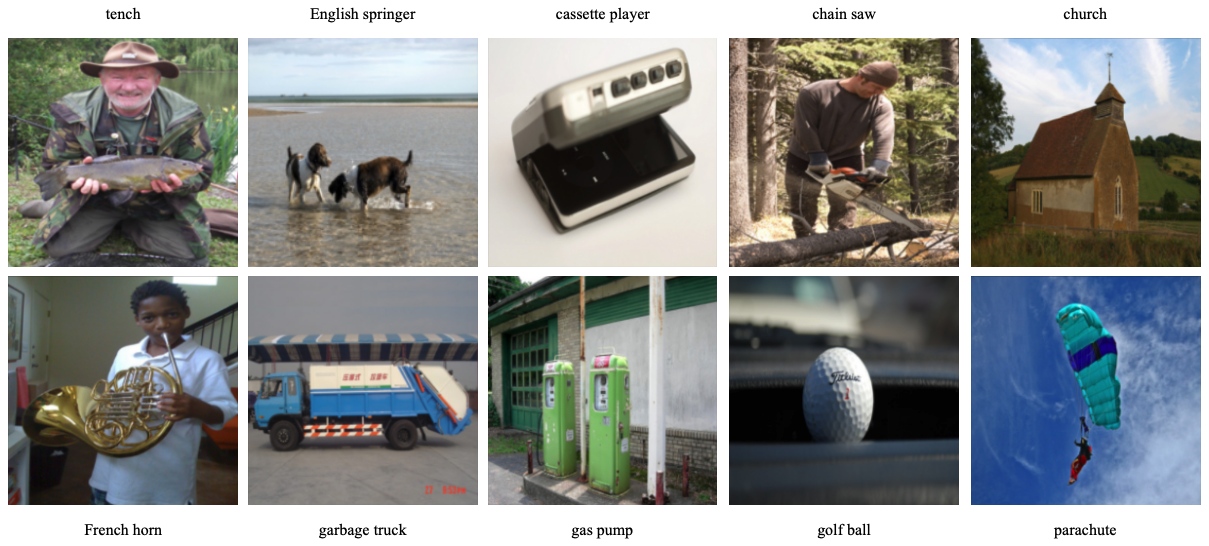}}
    \caption{Example images of the Imagenette dataset with their labels.}
    \label{fig:imagenette}
\end{figure}

We choose the Imagenette\footnote{Online: \url{https://github.com/fastai/imagenette} [accessed: \today]} dataset by FastAI for all experiments. Imagenette is a subset of Imagenet~\citep{deng2009imagenet}, one of the largest image databases for model benchmarks and research, and consists of ten classes, namely \textit{tench, English springer, cassette player, chain saw, church, French horn, garbage truck, gas pump, golf ball,} and \textit{parachute} \citep{imagenette}. Figure~\ref{fig:imagenette} shows one sample of each class in the dataset. The smaller size of the dataset requires fewer computational resources while still allowing the usage of models pre-trained on Imagenet as they share a similar data distribution. For the experiments, the class \textit{tench} was changed to \textit{fish}, and the class \textit{English springer} was changed to \textit{dog}. These changes simplify the Imagenette classes, as we could not anticipate our participants to have knowledge about fish species or dog breeds.

We employ the VGG-16 (configuration \textit{D}) model proposed by \citet{simonyan2014very}. 
Its structure is simple and homogeneous, consisting of 13 convolutional layers and 3 fully connected layers. Due to its depth and high number of parameters (138 million) VGG-16 provides an adequate architecture for applying NN pruning.
We initialize the model with pre-trained weights and fine-tune the model on Imagenette, resulting in a training accuracy of 99\%, a validation accuracy of 98\%, and a top-1 accuracy of 97\% on the testset.

\subsubsection{Pruning Approach}
\label{sususec:pruneapproach}
\begin{figure*}[t]
    \centering
    \begin{subfigure}[b]{.3\textwidth}
        \centering
        \includegraphics[width=.85\linewidth]{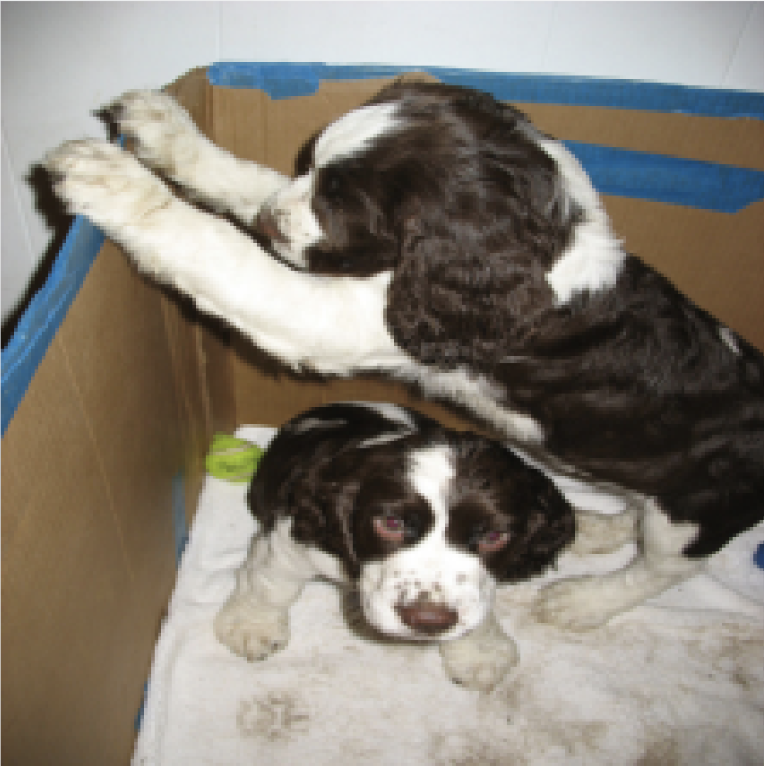}
        \caption{Original Image}
    \end{subfigure}
    \begin{subfigure}[b]{.3\textwidth}
        \centering
        \includegraphics[width=.85\linewidth]{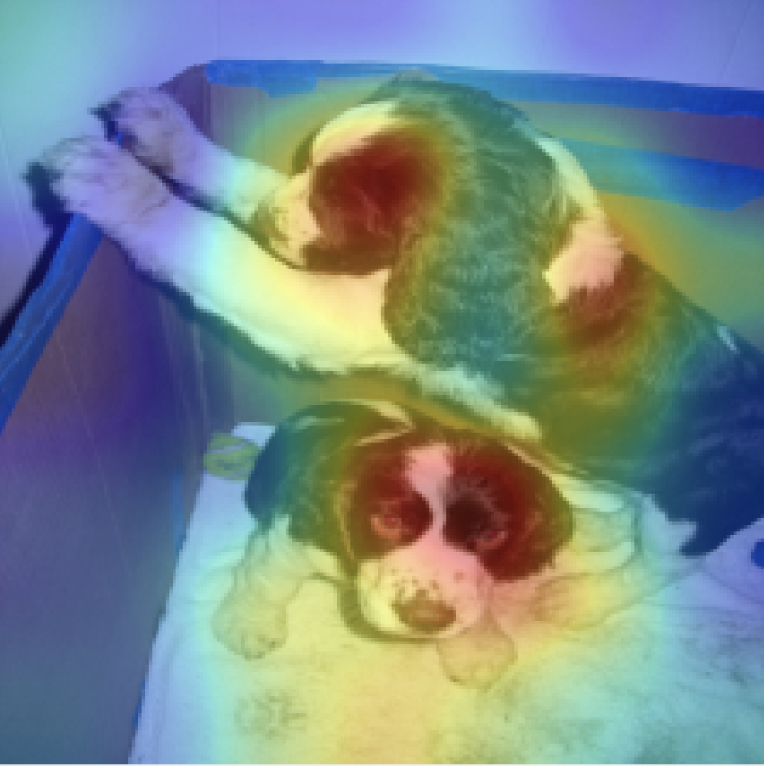}
        \caption{Heat map}
    \end{subfigure}
    \begin{subfigure}[b]{.3\textwidth}
        \centering
        \includegraphics[width=.85\linewidth]{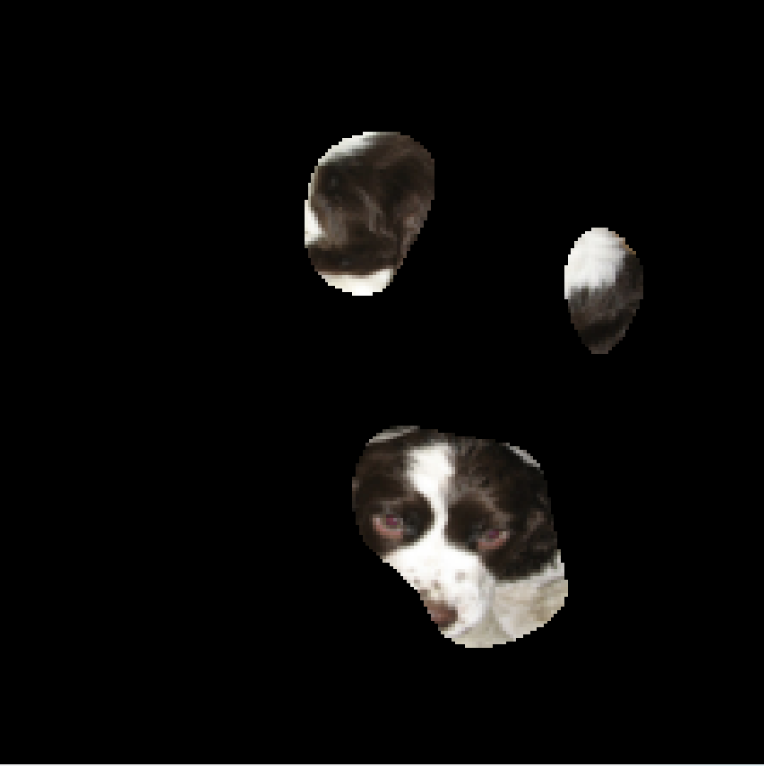}
        \caption{Occlusion map}
    \end{subfigure}
\caption{An image of the class 'dog', its heat- and occlusion map based on Grad-CAM and our calculations}
\label{fig:dog-heat-occ-example}
\end{figure*}

Following common baseline conventions in NN pruning research (e.g., \citep{han2015learning}, \citep{gale2019state}), we use iterative magnitude-based weight pruning. According to the five dimensions introduced in Sec.~\ref{susec:nnpruning}, this means, for the \textit{selection criterion} we chose the absolute values of the parameters, for the \textit{scope} we opt for local, i.e., layer-wise pruning, for the \textit{scheduling}, we follow the iterative paradigm, our \textit{pruning-structure} is unstructured, and for \textit{fine tuning}, we retrain every pruned model, for every CR \{2, 4, 8, 16, 32\}, after the pruning is completed until convergence, using the remaining weights with their pre-pruning values.
Although other choices are possible, magnitude-based pruning methods are a reliable choice as they are widely adopted in current research and have been proven to yield competitive results in comparison with more sophisticated approaches~\citep{blalock2020state}. Furthermore, previous research~\citep{han2015learning} has shown that magnitude-based methods yield better results when conducted iteratively. Thus, we refrain from examining one-shot pruning and implement our pruning methods strictly with an iterative pruning schedule. 
To ensure reproducibility, we utilize the PyTorch pruning library~\citep{paganini2020streamlining}. Additionally, we implement an object-oriented structure that allows us to easily and safely prune the convolutional layers of our converged VGG-16.

Furthermore, as convolutional layers contribute most to the inference time of CNNs and thus, have the greatest potential for theoretical speed-ups~\citep{molchanov2016pruning}, we only prune the convolutional layers of the network. 
Figure~\ref{fig:pruningacc} (on page~\pageref{fig:pruningacc}) shows the top-1 accuracy on the Imagenette test-set for every CR, including the accuracy of the unpruned VGG-16 (= CR 1) (in dark blue).

\subsubsection{Explainability Method}
\label{sususec:explainmeth}
As mentioned in Sec.~\ref{sec:xAImethods}, backpropagation-based techniques visualize areas of high network sensitivity and high network activation~\citep{gilpin2018explaining}, while only requiring one or few forward and backward passes~\citep{ancona2017towards}. Thus, we opt for the Grad-CAM approach due to its objective qualification as an explainability-method~\citep{adebayo2018sanity} and its relevance in both research and real-world applications.

Gradient-weighted Class Activation Mapping (Grad-CAM), proposed by \citet{selvaraju2020grad}, is one of the most prominent backpropagation-based attribution methods. Grad-CAM applies post-hoc to any CNN-based architecture and enables class-discriminative visualization.
As shown by \citet{mahendran2015natural} and \citet{bengio2012feature}, very deep convolutional layers capture instance-specific information and different types of image structures. Spatial information is then lost in the following fully connected layers. Therefore, deep convolutional layers represent the best option to capture higher-level semantics and spatial information. Grad-CAM aims to capture these high-level semantics and detailed spatial information from the network in order to identify image parts that were important for the classification decision. Similar to CAM, Grad-CAM uses the feature maps of the last convolutional layer. 
To calculate the Grad-Cam heat map, first, the gradient of the score for the respective image class $y^c$ is calculated with respect to the activation map (outputs) $A$ of the chosen target convolutional layer (commonly the last convolutional layer), i.e., $\frac{\partial y^{c}}{\partial A_{ij}^{k}}$. Global average pooling then yields a vector $\alpha_k^c$ with a weight for each channel of the activation map.
\[
    \alpha_{k}^{c} = \overbrace{\frac{1}{Z} \sum_{i}\sum_{j}}^{\text{global~average~pooling}} \frac{\partial y^{c}}{\partial A_{ij}^{k}}
\]

Next, the activation map $A^k$ is multiplied with the weight vector $\alpha_k^c$ and all channels are summed up, producing a heat map with the same height and width as the convolutional layer output. Finally, a $ReLU$ operation is performed, canceling out all below-zero values. Similar to CAM, also Grad-CAM up-samples and normalizes the resulting heat maps for visualization.
\[
    L_{\text{GradCAM}}^{c} = ReLU \left( \sum_{k} \alpha_{k}^{c}  A^{k} \right)
\]

We implement Grad-CAM, utilizing the PyTorch hooking mechanisms 
and craft two types of images from the resulting activation-map matrices: Heat maps and occlusion maps. Figure \ref{fig:dog-heat-occ-example} includes an example image with its heat map and occlusion map.
For the \textbf{heat map} images, the activation-map matrix is up-sampled to the input image size of $(224 \times 224)$ pixels and visualized in colors from red (very important) to blue (least important). We then apply the heat map over the original image with an opacity of $\alpha=0.4$.
For the \textbf{occlusion map} images, the activation-map matrix is up-sampled to the input image size. Instead of colors indicating the importance, the least important 90\% of the pixels are masked. These occlusion maps are inspired by the visualizations in Grad-CAM++\citep{chattopadhay2018gradcam++}, where the idea is to display only the parts of the image that appear to be most important to the network's decision. After visual inspection of some examples with varying degrees of occlusion, we employ an occlusion rate of 90\%.

\subsection{Experimental Setup}
\label{susec:expsetup}

Our research design is based on a three-phase approach, containing a pre-study and two more focused experiments. While the pre-study and the first experiment explore the participants' subjective perception, the second experiment evaluates the participants' performance based on objective metrics, leveraging \citet{schmidt2019quantifying} findings that more accurate decision-making indicates better explanation.

We used Amazon's Mechanical Turk\footnotemark[1] (MTurk) platform for all experiments to gather sufficient data. 
To ensure high-quality results, only MTurk respondents with a human intelligence task approval rate greater than 90\% were admitted.

In the following, we present the setups for each of these phases and rationalize their design choices. 
\subsubsection{Pre-study}
For the pre-study, we create heat- and occlusion maps for the models with CR 1, 2, and 32. 
CR 1, i.\,e., the non-pruned model, acts as the baseline. The great distance between CR 2 and CR 32 provided results from two different ends of the pruning spectrum. We use 500 Imagenette test-set images -- 50 of each class. All test-set images are predicted correctly by all three models. For every image and CR, we created a heat- and an occlusion map, resulting in three heat maps and three occlusion maps per image, i.e., a total of 1\,500 heat maps and 1\,500 occlusion maps. Participants were confronted with three images: 
\begin{enumerate}
\item the original image, always placed in the middle with the correctly predicted class,
\item a heat map image crafted with the unpruned model, randomly placed on either the left or right side, and
\item a heat map image crafted by either the CR 2 model or the CR 32 model on the other side.
\end{enumerate}
The setup generates a total of 1\,000 unique tasks (two for each original image).
The CR is unknown to the participants, assuring a blind study setup. Participants were asked to select the algorithm whose predictions they believed were more reasonable, or in case both algorithms felt equally reasonable select the middle point, effectively representing a three-point Likert scale. 
\fm{We have deliberately chosen the less-technical term 'reasonable' as initial tests have shown that participants had a better understanding of this term than 'explainable'.}
The same setup was used for the occlusion map images, resulting in another 1\,000 unique tasks. Exemplary setups of the pre-study can be found in Figures~\ref{fig:expsetup-heat} and~\ref{fig:expsetup-occ} in Appendix~\ref{sec:apppre}.
Each of the 2\,000 unique tasks was answered five times, generating a total of 10\,000 answers.

\subsubsection{Experiment 1}
\label{sec:method:exp1}
Lessons learned from our pre-study, that influenced the setup of Experiment 1 were: First, the CRs do influence the explainability of the CNN, second, heat maps seem to be more suitable than occlusion maps for the evaluation with a subjective metric, and third, CR 32 has a negative impact on the accuracy (cf. Figure~\ref{fig:pruningacc}, on page~\pageref{fig:pruningacc}) and the explainability to humans\footnote{We report detailed results of the pre-study in Section~\ref{susec:prestudy}.}. 
With these lessons in mind, we use heat maps only for Experiment 1. Furthermore, we extend the three-point Likert scale to a five-point Likert scale and disregard the model with CR 32 in favor of models with CR 4 and CR 8.
Figure~\ref{fig:exsetup} (on page~\pageref{fig:exsetup}) shows the presentation of a single task.
For this experiment, we focus on the heat map images, as the pre-study has shown that these images produce clearer results.
Again, participants were asked to decide which of the algorithms was more reasonable in its decision-making based on the shown heat map images. This time, participants were asked to make a more nuanced selection based on a 5-point Likert scale running from \textit{clearly more reasonable} (cmr) to \textit{clearly less reasonable} (clr) with \textit{slightly more reasonable} (smr), \textit{equally reasonable} (eq), and \textit{slightly less reasonable} (slr) in between.
We crafted 500 heat map images for the CRs ${1,2,4,8}$, and compared each CR with every other CR, resulting in 3\,000 unique tasks. Every unique task was answered by 5 respondents, generating a total of 15\,000 answers.

\vspace{0.5cm}
\noindent\textbf{Evaluation}\\
To calculate inter-coder agreement, we encode all answer possibilities with values ranging from -2 for \textit{clearly less reasonable} to +2 for \textit{clearly more reasonable}. We then calculate Krippendorff's $\alpha$~\citep{krippendorff2011computing} to evaluate the agreement level.

\begin{table}[tb]
    \centering
    \begin{tabular}{ccccc@{\hskip 0.5cm}c}
    	\rot{cmr} & \rot{smr} & \rot{eq} & \rot{slr} & \rot{clr} & \\
        ~[+2,&+1,&0,&-1,&-2]& standard deviation\\
        \toprule
         ~[0,&0,&0,&0,&5] & 0 \\
         ~[0,&0,&0,&1,&4] & 0.4 \\
         ~[0,&1,&1,&2,&1] & 1.0198\\
         ~[0,&1,&1,&1,&2] & 1.1662\\
         ~[1,&1,&1,&1,&1] & $\sqrt{2}$ \\
         ~[1,&1,&0,&2,&1] & 1.4697 \\
         ~[2,&0,&0,&0,&3] & 1.9596 \\
    \end{tabular}
    \caption{Examples of inter-rater agreement and the corresponding standard deviations.}
    \label{tab:stds}
\end{table}

Additionally, we report the inter-rater agreement based on the standard deviation of the respondents per task. With five respondents per task and the answers encoded as mentioned above, the standard deviation may compute to 26 different values ranging from 0 (where we have full agreement, i.e., all respondents choose the same option) to 1.9596 (where we have complete disagreement, e.g., two respondents answer with clearly more reasonable and three with clearly less reasonable or the other way around). 
Thus, we can say that lower values indicate more agreement among respondents, while higher values translate to more disagreement. The value of $\sqrt{2}$ (i.e., one answer per option) may serve as a baseline that indicates randomness, meaning that values below $\sqrt{2}$ tend to represent more agreement and values above more disagreement.
To give an intuition we sampled some of these 26 possibilities in Table~\ref{tab:stds}.

As we cannot assume explainability to be transitive (i.e., even if the CR 2 model produces more reasonable heat maps than the CR 1 model and the CR 4 model produces more reasonable heat maps than the CR 2 model, we cannot be sure that the CR 4 model is more reasonable than the CR 1 model), we observe all tasks that contain a specific CR.
Exemplary for CR 1, we accumulate all tasks that compare CR 1 with either CR 2, CR 4, or CR 8. The mean of all respondents' answers, encoded as above, creates a metric that describes the explainability of this specific model relative to all other examined models. We define this number as a model's \textit{explainability index} 
which allows for a direct comparison between all algorithms. A higher 
value indicates superior explainability compared to the remaining algorithms.

\subsubsection{Experiment 2}
\begin{figure}[tb]
    \graphicspath{{./figures/}}
    \centering
    \resizebox{.5\linewidth}{!}{\includegraphics{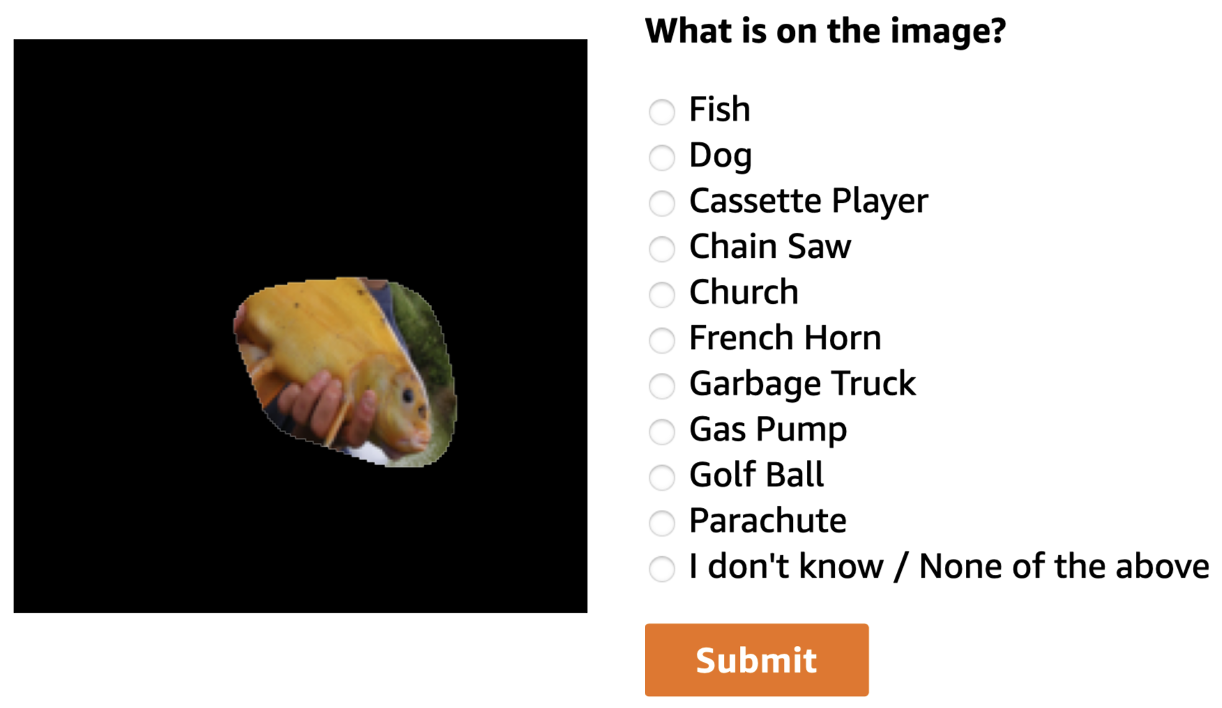}}
    \caption{Experimental Setup - Experiment 2: occlusion map (CR 1).}
    \label{fig:expsetup-exp2}
\end{figure}

Experiment 2 aims to assess the effects of different CRs on Grad-CAM occlusion maps and how they differ from human-understandable areas. 
Participants were confronted with a single occlusion map, crafted as described in Section \ref{sec:technical_setup}, i.e., based on models with CR $1, 2, 4, 8$, and $32$. 
All occlusion maps appear in random order to prevent the occlusion map of an identical image. 
Next to the occlusion map, all ten Imagenette classes are listed as possible answers, as depicted in~Figure~\ref{fig:expsetup-exp2}. The respondents were instructed to choose the most suitable class for the map. In case they felt that none of the classes would fit, they were asked to choose `I don't know / None of the above'. The accuracy of the respondents' answers provides an objective evaluation metric.
Note that this setup relies on occlusion instead of heat maps as participants would be able to see the whole image and thus know exactly what the image displays.
%
In total, this results in 2\,500 unique tasks, as 5 occlusion maps of 500 images were crafted - one for every CR. Every unique task was answered five times, generating a total of 7\,500 answers. 

\vspace{0.5cm}
\noindent\textbf{Evaluation}\\
To assess the inter-rater agreement we calculate Krippendorff's $\alpha$~\citep{krippendorff2011computing}. As for this experiment, a clear ground truth is available we are able to calculate the human-observer accuracy as the ratio of correct answers and the total amount of answers. Additionally, we report the error rate as the ratio between wrong answers and the total number of answers as well as the ratio of images for which the respondents are indecisive. We report these numbers for each CR in $\{1,2,4,8,32\}$ over all classes and for each individual class.

\begin{table}[tb]
\centering
\resizebox{.5\linewidth}{!}{
\begin{tabular}{lccc}
\toprule
\multicolumn{1}{l}{} &
  \multicolumn{1}{c}{\begin{tabular}[c]{@{}c@{}}\textbf{CR 1 vs CR 2}\\ \textit{(more reasonable in \%*)}\end{tabular}} &
  \multicolumn{1}{c}{\begin{tabular}[c]{@{}c@{}}\textbf{CR 1 vs CR 32}\\ \textit{(more reasonable in \%*)}\end{tabular}} &
  \\ 
\multicolumn{1}{l}{\textbf{heat maps}}        & \multicolumn{1}{c}{46.6\% vs 53.4\%} & \multicolumn{1}{c}{52.9\% vs 47.1\%} & 
\\ 
\multicolumn{1}{l}{\textbf{occlusion maps}} & \multicolumn{1}{c}{45.0\% vs 55.0\%} & \multicolumn{1}{c}{54.2\% vs 45.8\%} & 
\\[1.5ex]
  & \textit{equally reasonable in \%} &
  \textit{equally reasonable in \%}\\ 
  \midrule
\multicolumn{1}{l}{\textbf{heat maps}} &
  38.68\% &
  13.16\% &
  \\ 
\multicolumn{1}{l}{\textbf{occlusion maps}} &
  53.72\% &
  26.40\% &
  \\ 
\bottomrule
\multicolumn{4}{r}{\footnotesize{* \% based on answers without ``equal reasonability''}}
\end{tabular}
}
\caption{Explainability comparison of pre-study results for different CRs and visualization methods.}
\label{tbl:CR-percentages}
\end{table}

\section{Results}
\label{sec:results}
Following we elaborate on the results of our three studies and reflect on their findings.
\subsection{Pre-study}
\label{susec:prestudy}
Investigating agreement levels, the pre-study setup with heat maps achieves a Krippendorff's $\alpha$ score of 0.13, while the setup with occlusion maps achieves a score of 0.24. This indicates a rather low agreement across respondents as to which CR for particular images appears more reasonable. The higher value for the occlusion maps stems from the fact that here more respondents believe that both models seem equally reasonable.

Furthermore, we find that the three CRs used (1, 2, and 32) do make a difference in explainability. 
The upper half of Table~\ref{tbl:CR-percentages} compares the percentages of explainability between heat- and occlusion maps. The results indicate that the unpruned model has worse explainability than the CR 2 model but is more explainable than the CR 32 model.

\begin{table}[tb]
    \centering
    \resizebox{.5\linewidth}{!}{ 
    \begin{tabular}{lcccccc}\toprule
    {} & total & CR 1 & CR 2 & CR 4 & CR 8 & \\ \midrule 
    Krippendorff's $\alpha$ & 0.086 & 0.0916 & 0.0777 & 0.0913 & 0.0904 & \\[1ex]
    mean over & \multirow{2}{*}{1.0036} & \multirow{2}{*}{0.9904} & \multirow{2}{*}{0.9784} & \multirow{2}{*}{1.0034} & \multirow{2}{*}{1.0424} & \\ 
    standard deviations &  &  &  &  &  & \\ \midrule
    & CR 1 vs. & CR 1 vs. & CR 1 vs. & CR 2 vs. & CR 2 vs. & CR 4 vs. \\ 
    & \multicolumn{1}{l}{CR 2} & \multicolumn{1}{l}{CR 4} & \multicolumn{1}{l}{CR 8} & \multicolumn{1}{l}{CR 4} & \multicolumn{1}{l}{CR 8} & \multicolumn{1}{l}{CR 8} \\ \midrule
    Krippendorff's $\alpha$ & 0.049 & 0.1168 & 0.0985 & 0.0806 & 0.0962 & 0.0752 \\[1ex]
    mean over & \multirow{2}{*}{0.8873} & \multirow{2}{*}{1.0185} & \multirow{2}{*}{1.0654} & \multirow{2}{*}{0.9889} & \multirow{2}{*}{1.0588} & \multirow{2}{*}{1.0029} \\ 
    standard deviations &  &  &  &  &  & \\ \bottomrule
    \end{tabular}
    }
    \caption{Inter-rater agreement for the first experiment. The ``total'' column is measured over all responses, the other four columns in the top row are calculated per CR against all other CRs, and all lower columns are calculated for the specified CR comparison.}
    \label{tab:exp1-inter-rater-agreement}
\end{table}

The lower half of Table~\ref{tbl:CR-percentages} presents the percentage of respondents that chose ``Both are equally reasonable''. We can see that this number is clearly higher for occlusion maps. A higher number for ``equally reasonable'' indicates that NN pruning does not have as much of an effect on the explainability of the occlusion maps.

Overall, Table~\ref{tbl:CR-percentages} suggests that occlusion maps are more robust to different pruning ratios, as can be seen by the higher number of indecisive respondents, and thus not as suitable to assess the difference between different pruning ratios as heat maps. Hence, one of the most important results from our pre-study is that considering the task of assessing explainability to humans, heat maps are more suitable than occlusion maps.

\subsection{Experiment 1: Which Algorithm is More Reasonable?}
\label{susec:experiment1}

We report the inter-rater agreement for the first experiment in Table~\ref{tab:exp1-inter-rater-agreement}. Independent of the applied CR, Krippendorf's $\alpha$ is relatively low, indicating low agreement among the respondents. As described in Section~\ref{sec:method:exp1}, we further report the mean over the per-task standard deviation. Over all tasks, we observe values between 0 (complete agreement) and 1.9596 (complete disagreement). 
Further, the mean of the standard deviation over all tasks is 1.0036, which is clearly below 
$\sqrt{2}$ (the standard deviation of a uniform distribution), indicating a certain level of agreement among the respondents.

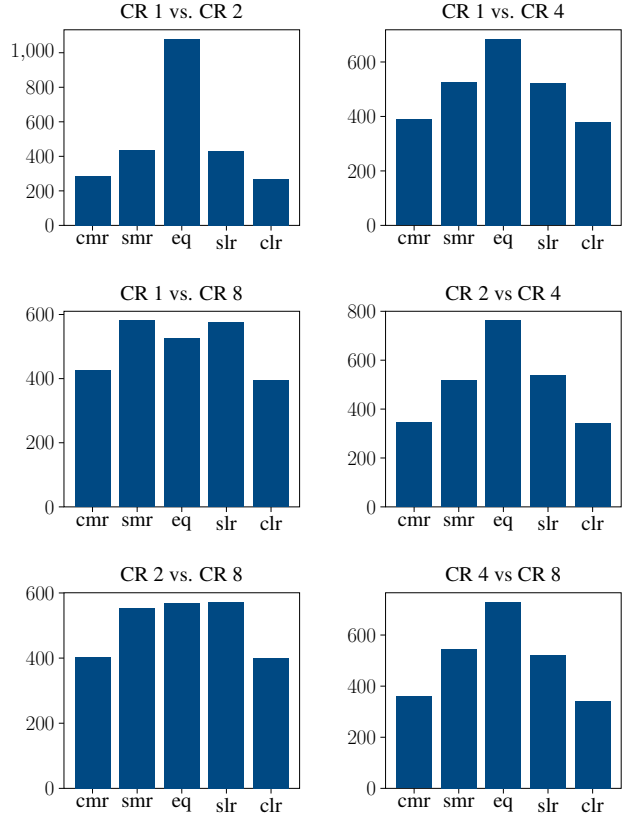
\begin{figure}[t]
	\centering
	\resizebox{.5\linewidth}{!}{%
\begin{tikzpicture}[font=\LARGE]

\definecolor{color0}{RGB}{0, 73, 131} 

\begin{groupplot}[group style={group size=2 by 3, horizontal sep=2.5cm, vertical sep=2.5 cm}]
\nextgroupplot[
tick align=outside,
tick pos=left,
title={CR 1 vs. CR 2},
x grid style={white!69.0196078431373!black},
xmin=-0.64, xmax=4.64,
xticklabels = {~,cmr,smr,eq,slr,clr},
xtick style={color=black},
y grid style={white!69.0196078431373!black},
ymin=0, ymax=1132.95,
ytick style={color=black}
]
\draw[draw=none,fill=color0] (axis cs:-0.4,0) rectangle (axis cs:0.4,284);
\draw[draw=none,fill=color0] (axis cs:0.6,0) rectangle (axis cs:1.4,435);
\draw[draw=none,fill=color0] (axis cs:1.6,0) rectangle (axis cs:2.4,1079);
\draw[draw=none,fill=color0] (axis cs:2.6,0) rectangle (axis cs:3.4,433);
\draw[draw=none,fill=color0] (axis cs:3.6,0) rectangle (axis cs:4.4,269);

\nextgroupplot[
tick align=outside,
tick pos=left,
title={CR 1 vs. CR 4},
x grid style={white!69.0196078431373!black},
xmin=-0.64, xmax=4.64,
xticklabels = {~,cmr,smr,eq,slr,clr},
xtick style={color=black},
y grid style={white!69.0196078431373!black},
ymin=0, ymax=718.2,
ytick style={color=black}
]
\draw[draw=none,fill=color0] (axis cs:-0.4,0) rectangle (axis cs:0.4,389);
\draw[draw=none,fill=color0] (axis cs:0.6,0) rectangle (axis cs:1.4,527);
\draw[draw=none,fill=color0] (axis cs:1.6,0) rectangle (axis cs:2.4,684);
\draw[draw=none,fill=color0] (axis cs:2.6,0) rectangle (axis cs:3.4,521);
\draw[draw=none,fill=color0] (axis cs:3.6,0) rectangle (axis cs:4.4,379);

\nextgroupplot[
tick align=outside,
tick pos=left,
title={CR 1 vs. CR 8},
x grid style={white!69.0196078431373!black},
xmin=-0.64, xmax=4.64,
xticklabels = {~,cmr,smr,eq,slr,clr},
xtick style={color=black},
y grid style={white!69.0196078431373!black},
ymin=0, ymax=610.05,
ytick style={color=black}
]
\draw[draw=none,fill=color0] (axis cs:-0.4,0) rectangle (axis cs:0.4,425);
\draw[draw=none,fill=color0] (axis cs:0.6,0) rectangle (axis cs:1.4,581);
\draw[draw=none,fill=color0] (axis cs:1.6,0) rectangle (axis cs:2.4,525);
\draw[draw=none,fill=color0] (axis cs:2.6,0) rectangle (axis cs:3.4,575);
\draw[draw=none,fill=color0] (axis cs:3.6,0) rectangle (axis cs:4.4,394);

\nextgroupplot[
tick align=outside,
tick pos=left,
title={CR 2 vs CR 4},
x grid style={white!69.0196078431373!black},
xmin=-0.64, xmax=4.64,
xticklabels = {~,cmr,smr,eq,slr,clr},
xtick style={color=black},
y grid style={white!69.0196078431373!black},
ymin=0, ymax=800.1,
ytick style={color=black}
]
\draw[draw=none,fill=color0] (axis cs:-0.4,0) rectangle (axis cs:0.4,345);
\draw[draw=none,fill=color0] (axis cs:0.6,0) rectangle (axis cs:1.4,517);
\draw[draw=none,fill=color0] (axis cs:1.6,0) rectangle (axis cs:2.4,762);
\draw[draw=none,fill=color0] (axis cs:2.6,0) rectangle (axis cs:3.4,536);
\draw[draw=none,fill=color0] (axis cs:3.6,0) rectangle (axis cs:4.4,340);

\nextgroupplot[
tick align=outside,
tick pos=left,
title={CR 2 vs. CR 8},
x grid style={white!69.0196078431373!black},
xmin=-0.64, xmax=4.64,
xticklabels = {~,cmr,smr,eq,slr,clr},
xtick style={color=black},
y grid style={white!69.0196078431373!black},
ymin=0, ymax=600.6,
ytick style={color=black}
]
\draw[draw=none,fill=color0] (axis cs:-0.4,0) rectangle (axis cs:0.4,404);
\draw[draw=none,fill=color0] (axis cs:0.6,0) rectangle (axis cs:1.4,554);
\draw[draw=none,fill=color0] (axis cs:1.6,0) rectangle (axis cs:2.4,569);
\draw[draw=none,fill=color0] (axis cs:2.6,0) rectangle (axis cs:3.4,572);
\draw[draw=none,fill=color0] (axis cs:3.6,0) rectangle (axis cs:4.4,401);

\nextgroupplot[
tick align=outside,
tick pos=left,
title={CR 4 vs CR 8},
x grid style={white!69.0196078431373!black},
xmin=-0.64, xmax=4.64,
xticklabels = {~,cmr,smr,eq,slr,clr},
xtick style={color=black},
y grid style={white!69.0196078431373!black},
ymin=0, ymax=765.45,
ytick style={color=black}
]
\draw[draw=none,fill=color0] (axis cs:-0.4,0) rectangle (axis cs:0.4,360);
\draw[draw=none,fill=color0] (axis cs:0.6,0) rectangle (axis cs:1.4,546);
\draw[draw=none,fill=color0] (axis cs:1.6,0) rectangle (axis cs:2.4,729);
\draw[draw=none,fill=color0] (axis cs:2.6,0) rectangle (axis cs:3.4,522);
\draw[draw=none,fill=color0] (axis cs:3.6,0) rectangle (axis cs:4.4,343);

\end{groupplot}

\end{tikzpicture}
	}
	\caption{The distribution of participants' answers for each model comparison. The higher CR is always compared to the lower CR. (Mind the different y-axes.)}
	\label{fig:CRs_baseline}
\end{figure}

Figure~\ref{fig:CRs_baseline} depicts all per comparison results. The first three graphs show the accumulated answers for the comparisons of the unpruned model with models pruned with CR 2, 4, and 8, respectively. We observe that the proportion of participants rating both algorithms equally reasonable declines when increasing the CR. Further, the last three graphs in Figure~\ref{fig:CRs_baseline} show the accumulated answers for the tasks in which two pruned models are compared. Apart from the change in the number of respondents that find both algorithms equally reasonable, visually no clear trends are detectable. However, in Table~\ref{tab:exp1-results} we report the mean of the respective encoded answers. Values above 0 indicate that the first-mentioned algorithm produces more explainable heat maps while values below 0 indicate the opposite. One can see that in total, pruning with CRs 2, 4, and 8 (upper three lines) seems to help the explainability to human raters. Transitivity, however, is not given. When considering the baseline experiments comparing the unpruned with the pruned models, we would expect the CR 8 model to produce better explainable heat maps than the CR 2 model. However, when comparing the respective heat maps directly, participants perceived the heat maps of the CR 2 model to be more reasonable than both, the CR 4 and CR 8 model (as can be seen in lines 4 and 5 of Table~\ref{tab:exp1-results}).

Table~\ref{tab:exp1-explainability-index} illustrates the explainability index as described in Section~\ref{sec:method:exp1}, that is the explainability of heat maps produced with every CR compared to every other CR. By this metric, measured over all tested images, heat maps produced with the CR 8 model explain the model's decision most reasonably. Interestingly, the CR 2 and the CR 4 models both produce more reasonable explanations than the unpruned model as the orange line in Figure~\ref{fig:pruningacc} (on p.~\pageref{fig:pruningacc}) indicates. 
\begin{table}[t]
    \centering
\begin{tabular}{ccr}
         Algorithm 1 & Algorithm 2 & Mean\\
         \toprule
         CR 1 & CR 2 & -0.0128 \\
         CR 1 & CR 4 & -0.0104\\
         CR 1 & CR 8 & -0.0272\\
         CR 2 & CR 4 & 0.0036\\
         CR 2 & CR 8 & 0.0048\\
         CR 4 & CR 8 & -0.0232\\
    \end{tabular}
    \caption{Mean over all answers for the specific comparisons with answers encoded from -2 (Algorithm 1 is clearly less reasonable) to +2 (Algorithm 1 is clearly more reasonable).}
    \label{tab:exp1-results}
\end{table}
    \begin{table}[t]
    \centering
\begin{tabular}{lr}
         Algorithm & Explainability Index\\
         \toprule
         CR 1 & -0.0168 \\
         CR 2 & 0.0071\\
         CR 4 & -0.0055\\
         CR 8 & 0.0152\\
    \end{tabular}
    \caption{Explainability index for all models
    }
    \label{tab:exp1-explainability-index}
\end{table}

\begin{table}[tbh]
    \centering
\begin{tabular}{lccc@{\hskip 24pt}r}\toprule
		{} &  Accuracy & Error-rate & Indecisive &Krippendorf's $\alpha$'s\\
		\midrule 
		CR 1 & 85.84\% & 4.36\% & 9.80\% & 0.799268 \\
		CR 2 & 86.40\% & 4.64\% & 8.96\% & 0.790250 \\
		CR 4 & 82.08\% & 5.36\% & 12.56\% & 0.744596 \\
		CR 8 & 81.20\% & 5.36\% & 13.44\% & 0.738209 \\
		CR 32 & 81.32\% & 4.48\% & 14.20\% & 0.744790 \\
		\bottomrule
	\end{tabular}
		\caption{Overall results and inter-rater agreement of Experiment 2.}
		\label{tab:exp2-overall-results}
\end{table}

\subsection{Experiment 2: What is on the Image?}
\label{susec:experiment2}

\begin{table*}[tb]
    \centering
    \scalebox{1}{
    \begin{tabular}{lrrrrr@{\hspace{.5cm}}rrrrr}\toprule
    	class & \multicolumn{5}{c}{Human rater accuracies} 										& \multicolumn{5}{c}{Human rater error rates}\\
    	\cmidrule(l{2pt}r{12pt}){2-6} \cmidrule(l{0pt}r{2pt}){7-11}
    {} &  CR 1 &  CR 2 &  CR 4 &  CR 8 &  CR 32 								&  CR 1 &  CR 2 &  CR 4 &  CR 8 &  CR 32 \\ \midrule 
    fish           &  91.2\% &  \textbf{92.4} \% &  84.0 \% &  77.6 \% &   76.4 \%				&   2.0\% &   \textbf{0.4\%} &   2.4\% &   4.0\% &    6.0\% \\
    dog            &  90.0\% &  95.2 \% &  95.6 \% &  \textbf{97.2} \% &   96.0 \%				&   1.2\% &  \textbf{ 0.8\%} &   2.0\% &   \textbf{0.8\%} &    \textbf{0.8\%} \\
    cassette player &  79.6\% &  \textbf{82.4} \% &  73.2 \% &  71.6 \% &   77.6 \%				&   7.2\% &   \textbf{5.6\%} &  10.8\% &  10.8\% &    8.4\% \\
    chainsaw       &  \textbf{66.0}\% &  \textbf{66.0} \% &  58.8 \% &  56.4 \% &   52.4 \%		&  \textbf{11.6\%} &  12.0\% &  14.8\% &  14.4\% &   12.0\% \\
    church         &  85.2\% &  \textbf{92.4} \% &  90.8 \% &  89.6 \% &   88.4 \%				&   2.0\% &   4.0\% &   1.6\% &  \textbf{0.8\%} &    1.6\% \\
    french horn     &  \textbf{94.0}\% &  \textbf{94.0} \% &  86.8 \% &  89.6 \% &   94.4 \%		&   2.0\% &   2.8\% &   6.0\% &   4.4\% &    \textbf{0.4\%} \\
    garbage truck   &  86.0\% &  \textbf{88.0} \% &  86.4 \% &  84.8 \% &   84.4 \%				&   6.8\% &   6.8\% &   6.0\% &   5.6\% &    \textbf{3.6\%} \\
    gas pump        &  \textbf{82.8}\% &  76.0 \% &  77.6 \% &  77.2 \% &   78.8 \%				&   \textbf{5.6\%} &   9.2\% &   8.0\% &   6.8\% &    8.4\% \\
    golfball       &  95.6\% &  \textbf{97.2} \% &  96.4 \% &  91.6 \% &   88.0 \%				&   2.0\% &   1.2\% &   \textbf{0.8\%} &   2.4\% &    2.0\% \\
    parachute      &  \textbf{88.0}\% &  80.4 \% &  71.2 \% &  76.4 \% &   76.8\% 				&   3.2\% &   3.6\% &   \textbf{1.2\%} &   3.6\% &    1.6\% \\ \midrule 
    total          & 85.85\% & \textbf{86.40}\% & 82.08\% & 81.20 \% &  81.32 \%				&    \textbf{4.36\%} &    4.64\% &    5.36\% & 5.36\%    &     4.48\% \\ \bottomrule 
    \end{tabular}
}
    \caption{Accuracies and error rates per class for human raters in Experiment 2. Maximum, resp. minimum are highlighted in boldface.}
    \label{tab:human-acc}
\end{table*}
We assess the inter-rater agreement for this experiment with Krippendorf's $\alpha$. Table~\ref{tab:exp2-overall-results} shows the $\alpha$-values over all answers and for the specific CRs. The overall $\alpha$ of 0.76 indicates a reasonably high agreement among the respondents. Interestingly, the inter-rater agreement is higher for the unpruned ($\approx$0.80) and CR 2 pruned ($\approx$0.79) models and declines for higher pruning ratios ($\approx$0.74).

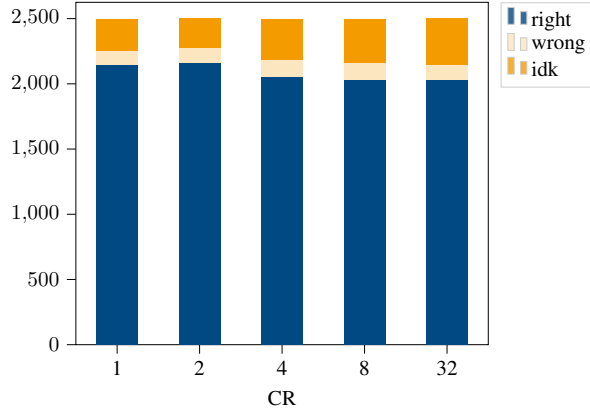
\begin{figure}[t]
 	\centering
 	\scalebox{.8}{
\begin{tikzpicture}

\definecolor{color0}{RGB}{0, 73, 131} 
\definecolor{color4}{RGB}{244, 155, 0} 
\colorlet{color1}{color0!75}
\colorlet{color2}{color4!25}
\colorlet{color3}{color4!75}

\begin{axis}[
legend pos=outer north east,
legend cell align=left,
legend cell align={left},
legend style={
  fill opacity=0.8,
  draw opacity=1,
  text opacity=1,
  draw=white!80!black
},
tick align=outside,
tick pos=left,
x grid style={white!69.0196078431373!black},
xlabel={CR},
xmin=-0.5, xmax=4.5,
xtick style={color=black},
xtick={0,1,2,3,4},
xticklabel style={rotate=0.0},
xticklabels={1,2,4,8,32},
y grid style={white!69.0196078431373!black},
ymin=0, ymax=2625,
ytick style={color=black}
]
\draw[draw=none,fill=color0] (axis cs:-0.25,0) rectangle (axis cs:0.25,2146);
\addlegendimage{ybar,ybar legend,draw=none,fill=color0}
\addlegendentry{right}

\draw[draw=none,fill=color0] (axis cs:0.75,0) rectangle (axis cs:1.25,2160);
\draw[draw=none,fill=color0] (axis cs:1.75,0) rectangle (axis cs:2.25,2052);
\draw[draw=none,fill=color0] (axis cs:2.75,0) rectangle (axis cs:3.25,2030);
\draw[draw=none,fill=color0] (axis cs:3.75,0) rectangle (axis cs:4.25,2033);

\draw[draw=none,fill=color2] (axis cs:-0.25,2146) rectangle (axis cs:0.25,2255);
\addlegendimage{ybar,ybar legend,draw=none,fill=color2}
\addlegendentry{wrong}
\draw[draw=none,fill=color2] (axis cs:0.75,2160) rectangle (axis cs:1.25,2276);
\draw[draw=none,fill=color2] (axis cs:1.75,2052) rectangle (axis cs:2.25,2186);
\draw[draw=none,fill=color2] (axis cs:2.75,2030) rectangle (axis cs:3.25,2164);
\draw[draw=none,fill=color2] (axis cs:3.75,2033) rectangle (axis cs:4.25,2145);

\draw[draw=none,fill=color4] (axis cs:-0.25,2255) rectangle (axis cs:0.25,2500);
\addlegendimage{ybar,ybar legend,draw=none,fill=color4}
\addlegendentry{idk}

\draw[draw=none,fill=color4] (axis cs:0.75,2276) rectangle (axis cs:1.25,2500);
\draw[draw=none,fill=color4] (axis cs:1.75,2186) rectangle (axis cs:2.25,2500);
\draw[draw=none,fill=color4] (axis cs:2.75,2164) rectangle (axis cs:3.25,2500);
\draw[draw=none,fill=color4] (axis cs:3.75,2145) rectangle (axis cs:4.25,2500);
\end{axis}

\end{tikzpicture}
 	}
 	\caption{Respondents' answers in Experiment 2 for all CRs.}
 	\label{fig:exp2-all-CRs}
 \end{figure}

Table~\ref{tab:exp2-overall-results} further shows the overall results of the second experiment. Mind that all values refer to the human-respondents' performance. We observe that the participants achieve a slightly higher accuracy for the occlusion maps produced with the CR 2 model (86.40\%) over those produced with the unpruned model (85.84\%). For higher pruning rates, the human-rater accuracy declined by 4.32\% (CR 4) to 5.2\% (CR 8). The human error rate is relatively stable, with the lowest value for the unpruned model (4.36\%) and the highest values for CR~4 and CR 8 (5.36\% each). For the ratio of answers in which the respondents have chosen the option `I don't know / None of the above' we observe a marginal decline from CR 1 (9.8\%) to CR 2 (8.96\%) and a subsequent rise up to 14.2\% for CR 32. 
Higher indecisiveness of the respondents indicates lower explainability due to indecipherable occlusion maps. However, indecisiveness might be preferable over a wrong answer since it would be better for an explanation map to explain nothing rather than explaining the wrong class.
The light blue line in Figure~\ref{fig:pruningacc} (on page~\pageref{fig:pruningacc}) shows the human rater accuracy and Figure~\ref{fig:exp2-all-CRs} illustrates the distribution of respondents' answers visually.

\begin{figure*}[t]
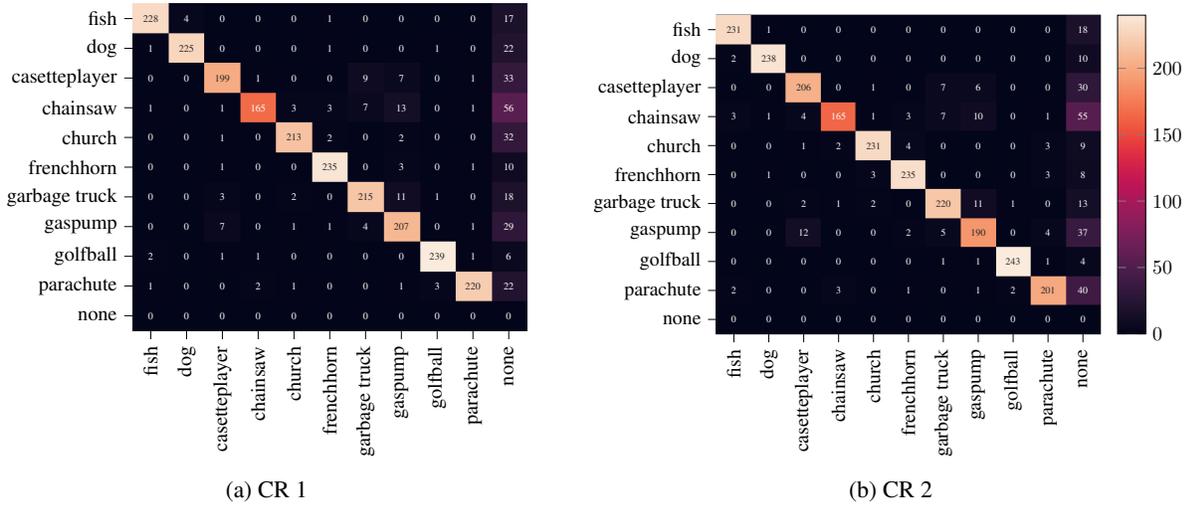

  \begin{subfigure}{.43\textwidth}
  \centering
  \resizebox{\linewidth}{!}{\input{fig/conf-matrix-CR1}}
  \caption{CR 1}
  \label{fig:confusion-matrix1}
  \end{subfigure}\qquad
  \begin{subfigure}{.49\textwidth}
  \centering
  \resizebox{\linewidth}{!}{\input{fig/conf-matrix-CR2}}
  \caption{CR 2}
  \label{fig:confusion-matrix2}
  \end{subfigure}
  \caption{Confusion matrices of the human ratings from Experiment 2 for all ten classes and CRs 1 (Subfigure \ref{fig:confusion-matrix1}) and 2 (Subfigure \ref{fig:confusion-matrix2}). Darker values indicate lower numbers and lighter values indicate higher numbers. The diagonals display correct classifications, while the right-most column shows the number of `I don't know / None of the above'. Confusion matrices of CR 4, 8, and 32 can be found in Appendix \ref{sec:appex2}.}
  \label{fig:confusion-matrices}
\end{figure*}

\begin{figure*}[t]
    \centering
    \begin{subfigure}[b]{.3\textwidth}
    \centering
    \includegraphics[width=.85\linewidth]{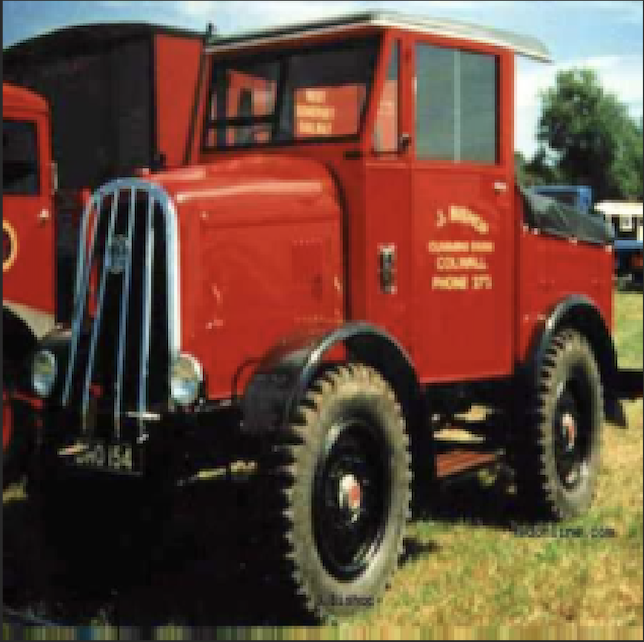}
    \caption{Original Image}
    \end{subfigure}
    \begin{subfigure}[b]{.3\textwidth}
    \centering
    \includegraphics[width=.85\linewidth]{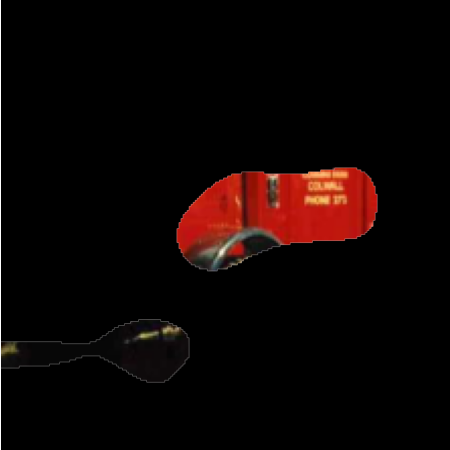}
    \caption{CR 1}
    \end{subfigure}
    \begin{subfigure}[b]{.3\textwidth}
    \centering
    \includegraphics[width=.85\linewidth]{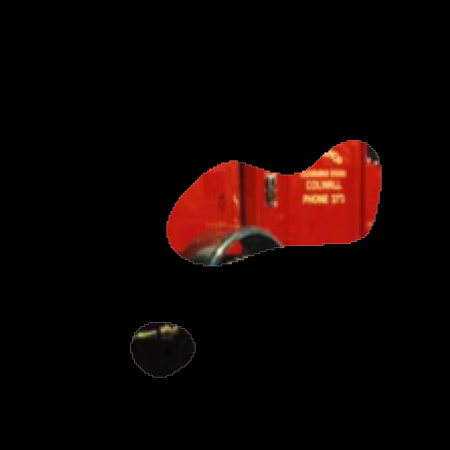}
    \caption{CR 2}
    \end{subfigure}

\caption{The original image and occlusion maps produced by the CR 1 and CR 2 models of the class `garbage truck'. 3/5 respondents misclassified the CR 1 occlusion maps as a gas pump, while 2/5 respondents misclassified the CR 2 model as a gas pump.}
\label{fig:truck-gaspump-misclass}
\end{figure*}

We find that the human-rater performance differs vastly between the classes. Table~\ref{tab:human-acc} displays the accuracies for all ten classes dependent on the CR of the used model.
For occlusion maps produced with the unpruned model, human-rater accuracy ranges from 66\% (chainsaw) to 95.6\% (golfball). Interestingly, the impact of NN pruning on the explainability of the produced occlusion maps is also dependent on the class. For some classes, the human-rater accuracy declines sharply when the underlying model is pruned. As an example, the human-rater performance for the class `parachute' is high (88\%) with the unpruned model, but heavily declines for CR 2 (80.4\%) and CR 4 (71.2\%). On the other hand, for the class `dog' respondents achieved the lowest accuracy (90\%) for occlusion maps produced with the unpruned model, while they were able to correctly classify the occlusion maps produced with the CR 8 model in more than 97\% of the cases.
A similar observation can be made when looking at the ratio of occlusion maps for which the respondents choose the option `None of the above / I don't know'. For the class `parachute' the number of indecisive respondents almost doubled when comparing occlusion maps from the CR 1 (8.8\%) and the CR 2 (16.0\%) model and rises even over 20\% for the CR 4, CR 8, and CR 32 model. Reversely, for the class `church' indecisiveness was the highest for the CR 1 model (12.8\%) while the CR 2 model produced the lowest value with 3.6\%. 
These findings suggest that the images' semantics impact the explainability of the models' decisions.
We include the full table for the respondents' indecisiveness in Appendix~\ref{sec:appex2}.

Error rates are relatively stable among the classes and the various models, as is visible in the right half of Table~\ref{tab:human-acc}. Respondents misclassified occlusion maps of the class `chainsaw' most often (11.6\%) and pruning slightly increases the error rate. For the class `cassette player', another class with a comparatively low human-rater accuracy (see Table~\ref{tab:human-acc}) and a higher error rate of 7.2\%, mild pruning (CR 2) decreases the error rate by 1.6\% to 5.6\%, while more extensive pruning increases the error rate to 10.8\% (CR 4 and CR 8).

The specific errors between the classes are visualized with confusion matrices in Figure~\ref{fig:confusion-matrices}. We observe higher error rates for occlusion maps of images containing chainsaws and garbage trucks. This holds true for CR 1 and CR 2 and is also visible for higher CRs. We provide the respective confusion matrices in Appendix~\ref{sec:appex2}. Figure~\ref{fig:truck-gaspump-misclass} illustrates a case in which fewer misclassifications occurred for the occlusion map produced by the pruned model.

\begin{figure*}[tbh]
    \centering
    \begin{subfigure}{\textwidth}
    \centering
    \includegraphics[width=\textwidth]{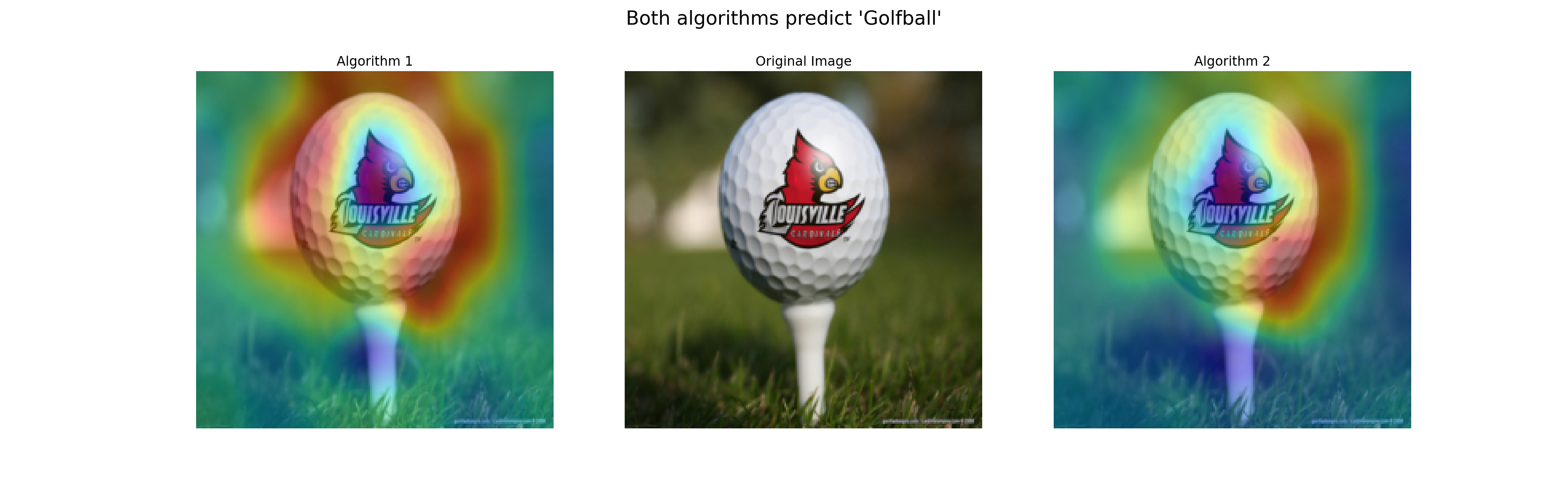}
    \caption{Heat maps produced with the CR 4 model on the left and the CR 8 model on the right. All participants marked algorithm 1 to be clearly more reasonable.}
    \label{fig:complete-agreement}
\end{subfigure}
\\~\\
\begin{subfigure}{\textwidth}
    \centering
    \includegraphics[width=\textwidth]{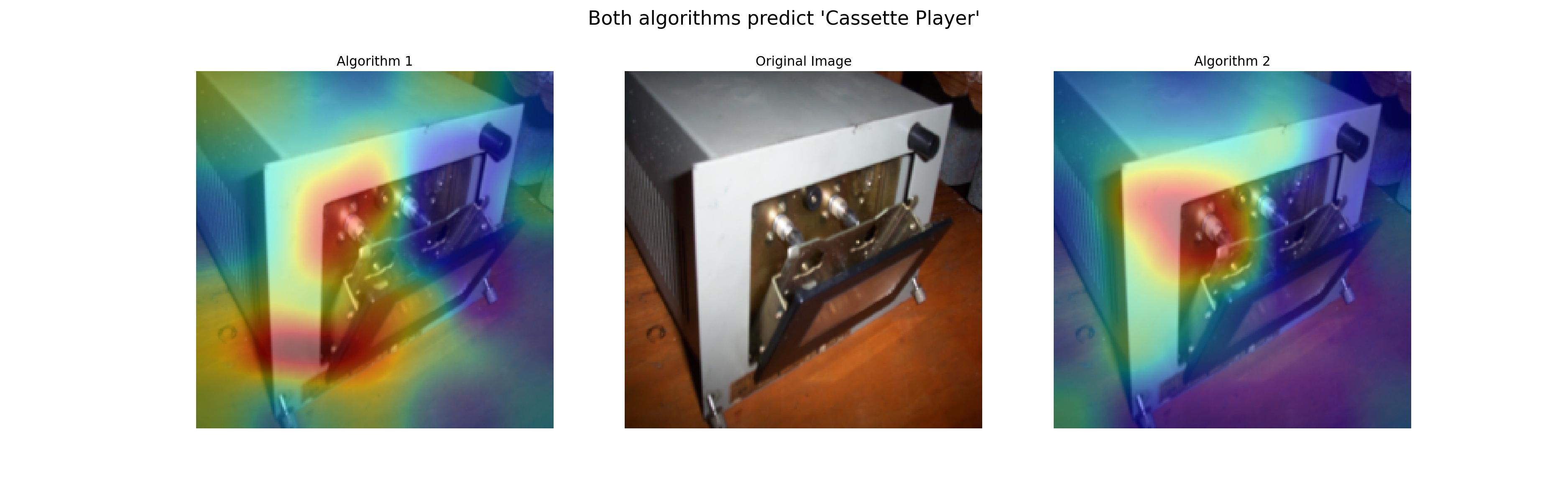}
    \caption{Heat maps produced with the CR 4 model on the left and the unpruned model on the right. Three participants marked algorithm 1 as clearly more reasonable, while two participants marked algorithm 2 as clearly more reasonable.}
    \label{fig:complete-disagreement}
\end{subfigure}
\caption{Samples from Experiment 1 that yield complete agreement and complete disagreement among the human raters}
\end{figure*}

\begin{figure*}[t]
    \centering
      \begin{subfigure}{.32\textwidth}
      \centering
      \includegraphics[width=.8\linewidth]{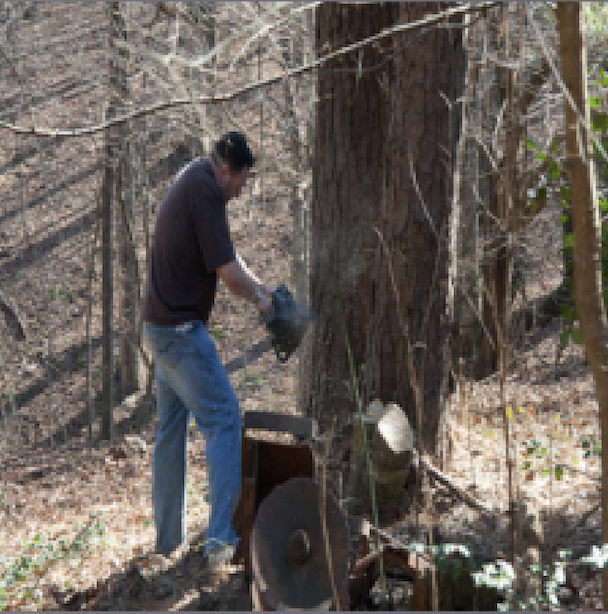}
      \caption{Original image}
      \end{subfigure}
      \begin{subfigure}{.32\textwidth}
      \centering
      \includegraphics[width=.8\linewidth]{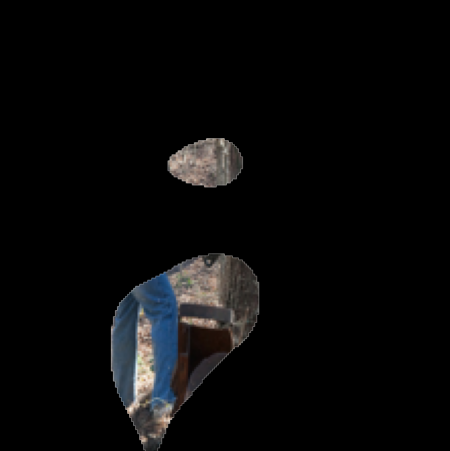}
      \caption{CR1}
      \end{subfigure}
      \begin{subfigure}{.32\textwidth}
      \centering
      \includegraphics[width=.8\linewidth]{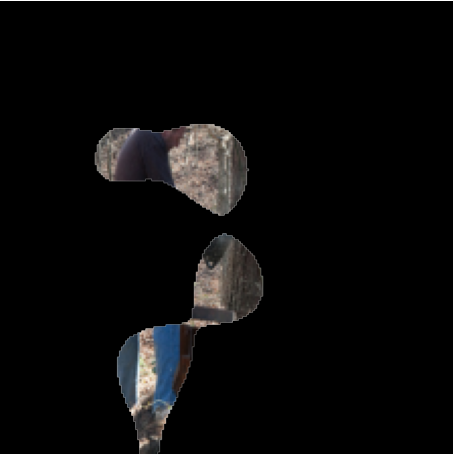}
      \caption{CR2}
      \end{subfigure}
      \\~\\
      \begin{subfigure}{.32\textwidth}
      \centering
      \includegraphics[width=.8\linewidth]{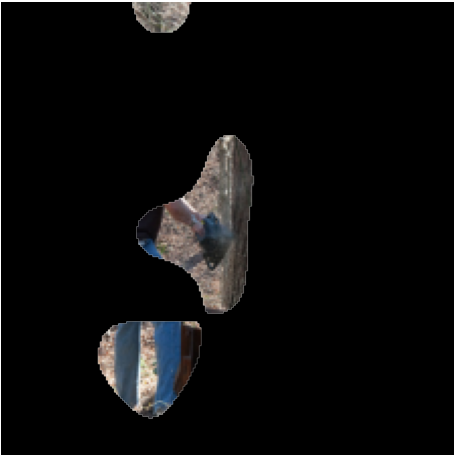}
      \caption{CR4}
      \end{subfigure}
      \begin{subfigure}{.32\textwidth}
      \centering
      \includegraphics[width=.8\linewidth]{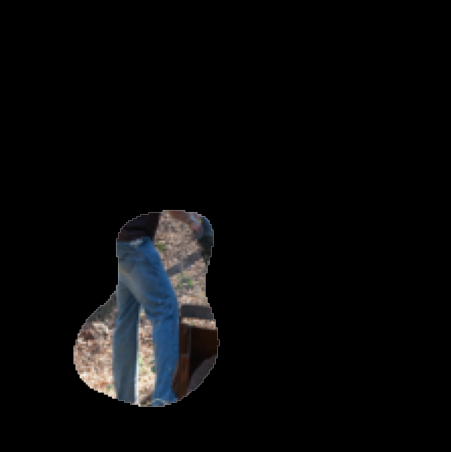}
      \caption{CR8}
      \end{subfigure}
      \begin{subfigure}{.32\textwidth}
      \centering
      \includegraphics[width=.8\linewidth]{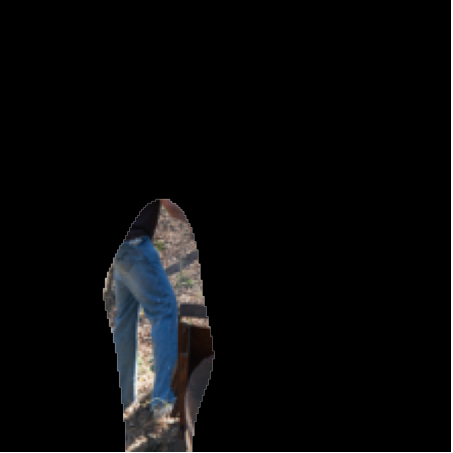}
      \caption{CR32}
      \end{subfigure}
      \caption{The original image and occlusion maps for one sample of the class chainsaw. Human rater accuracies are 1/5 for CR 1, 4/5 for CR 2, 5/5 for CR 4, and 0/5 for CR 8 and CR 32.}
      \label{fig:exp2-sample}
    \end{figure*}

\section{Limitations}
\label{sec:discussion}
We measure only low reliability scores for Krippendorff's $\alpha$ for the pre-study and the first experiment. We believe that these scores are caused by the following: As pointed out in~\citep{rudin2018please9}, the concept of explainability is subjective and domain-specific. Therefore, comparing the reasonability of two algorithms might be subjective to every respondent, especially when the two heat- or occlusion maps do not show significant differences. This may lead to different perceptions of the visualizations that differ among the respondents. It may further explain why the scores for Krippendorff's $\alpha$ are higher in the pre-study, where a three-point Likert scale was provided than in Experiment 1, where we provided a five-point Likert scale. 

Looking at specific samples might help in understanding how complete agreement or complete disagreement might emerge. Subfigure~\ref{fig:complete-agreement} illustrates one example in which all respondents agree that algorithm 1 is more reasonable. While the right heat map highlights some parts on the right of the ball, the left heat map covers the whole ball. Both heat maps assign less importance to the print in the middle of the ball. Additionally, the heat map on the right highlights the tee more concisely. Conversely, Subfigure~\ref{fig:complete-disagreement} illustrates the case of complete disagreement, i.e., three respondents chose algorithm 1 to be clearly more reasonable, and two respondents chose algorithm 2 to be clearly more reasonable. The model on the left seems to take more details into account. The participants' degree of familiarity with the subject might be an influencing factor regarding which algorithm appears to be more reasonable.
\fm{
The lack of a definite ground truth encountered in the first experiment highlights one of the most pressing issues within explainable AI research. \citet{muller2021kandinsky} proposed the Kandinsky Patterns as a potential solution. These abstract patterns have easily extractable structures of geometric shapes that might allow an evaluation against a given ground truth.}

In Experiment 2, we report a rather high reliability score. In comparison to the subjectiveness of perceived explainability, Experiment 2 proposes an objective measure with clear ground truth, as an image always includes one of the ten Imagenette classes. If the respondents were not able to recognize the image class, they had the chance to choose `I don't know / None of the above'.
Figure~\ref{fig:exp2-sample} shows the occlusion maps for all five CRs. For this sample, inter-rater agreement is reasonably high. Respondents who did not choose the correct answer all selected `I don't know / None of the above'. Looking at the occlusion maps, we observe that the CR 1, CR 8, and CR 32 models base their prediction more on latent features of the image such as the pose of the human and the presence of a tree, while the CR 2 and particularly the CR 4 model base their decision on the object in question, a chainsaw, itself. Hence, it does not seem surprising that humans struggle to classify these samples correctly.
\section{Future Work}
\label{sec:limitations}
Directions for future work, identified through our analysis, include the extension of the five-point Likert scale to a seven-point Likert scale and an experimental setup with more raters per task, maybe at the cost of a lower number of images. Carefully selecting these images, for example, by some kind of image complexity measure, would furthermore shed light on the open question about the relation between image complexity and explainability. Additionally, our setup is restricted to one explainability method and one pruning method, while there is a lot of ongoing work in each of these areas. 

Besides these direct extensions of our experiments, there are several dimensions in which our work can be diversified to reach more reliable results, especially regarding generalizability. First and foremost, the generalization to other CNN architectures such as 
ResNets \citep{he2016deep}, 
Inception \citep{szegedy2015going} or EfficientNets \citep{tan2019efficientnet} should be examined. Given that we choose an explainability method that is applicable to all CNNs, the open question is not if our methodology is applicable but rather if the results obtained for VGG-16 in this work also hold for other CNN architectures, and subsequently also for the novel class of transformer-based computer vision architectures, such as Swin \citep{liu2021Swin} and Vision Transformers \citep{Dosovitskiy2021AnII}.
\fm{
The development of a unified evaluation metric for explainability and network compression as proposed by \citet{yu2023x} is a promising direction to optimize both objectives simultaneously during the training process.}
Finally, it might be worthy to examine the impact of NN pruning on the internal mechanisms of GradCAM (e.g., in the used activation maps) and how these changes are reflected in our human-grounded experiment results.
Exploring each of these dimensions is a valuable direction for future work and given that we carefully selected our setup, we are positive that similar setups will result in similar sweet spots.
\section{Conclusion}
\label{sec:conclusion}
Our results suggest that there exists a sweet spot of mild pruning, that helps explainability without hurting human decision accuracy. This might not come as a surprise as a lower number of (important) parameters seems to intuitionally go along with higher explainability. But, our experiments also showed that for explainability research we always have to cover both, the subjective and the objective aspects. We did so by covering the subjective aspect in Experiment 1, which indicates that a CR of 8 produces the best explanations. Taking the objective measure of human decision accuracy in Experiment 2 into account, however, we see that all CRs above 2 have a negative influence on the accuracy of the participants. 

It is common knowledge in the machine learning community and can be seen in Figure~\ref{fig:pruningacc} (on p.~\pageref{fig:pruningacc}) that mild pruning also increases the accuracy of the underlying DNN~\citep{han2015learning}. Furthermore, turning to \textit{adversarial machine learning}, which is concerned with the security of machine learning algorithms, we find evidence that mild pruning increases the robustness of the classifiers against malicious adversaries~\citep{merkle2021pruning}.

Combining our results on NN pruning and explainability with these results from machine learning and adversarial machine learning suggests that NN pruning might be a ``jack of all trades'', decreasing complexity, computation time, and power consumption while simultaneously increasing explainability, accuracy, and security.

\bibliographystyle{unsrtnat}
\bibliography{references}  

\appendix

\newpage

\section{Additional Material from the Pre-study}
\label{sec:apppre}
\begin{figure}[h]
    \centering
    \resizebox{\linewidth}{!}{
    \includegraphics[]{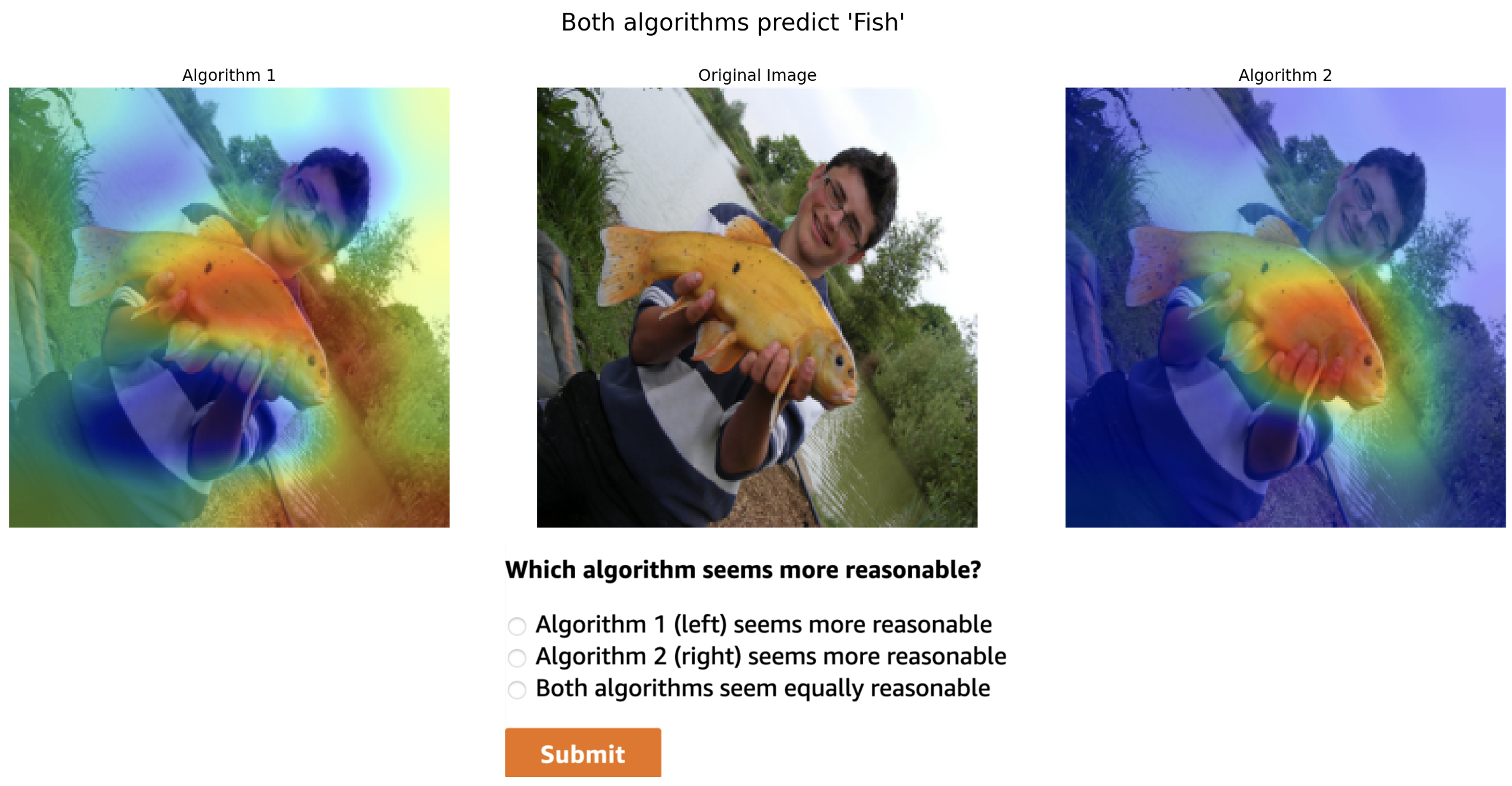}}
    \caption{\hbox{Experimental Pre-study setup with heat maps CR 32 (left) vs. CR 1 (right).}}
    \label{fig:expsetup-heat}
\end{figure}

\begin{figure}[htb]
    \centering
    \resizebox{\linewidth}{!}{
    \includegraphics[]{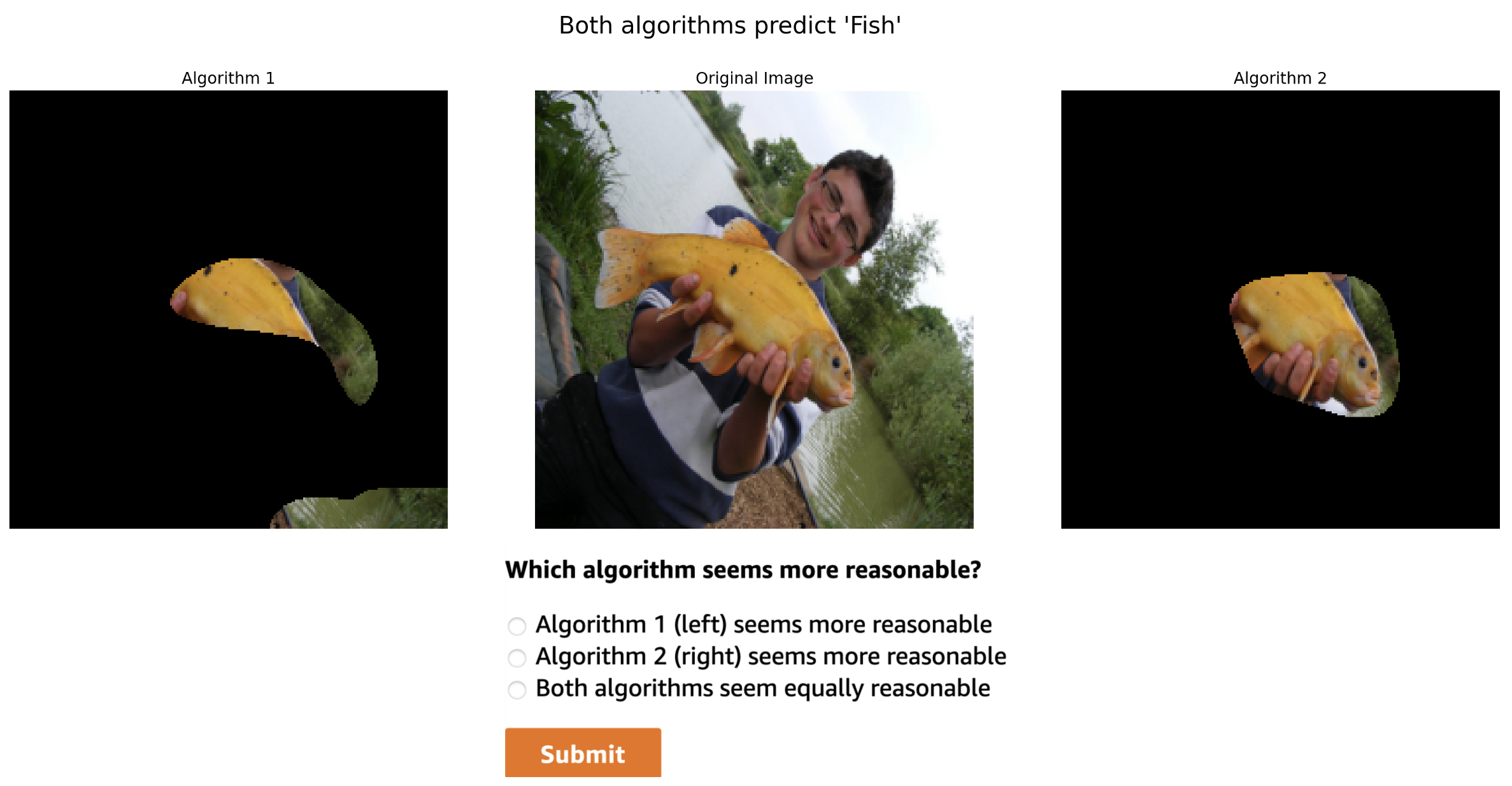}}
    \caption{\hbox{Experimental Pre-study setup with occlusion maps CR 32 (left) vs. CR 1 (right).}}
    \label{fig:expsetup-occ}
\end{figure}

Figures \ref{fig:expsetup-heat} and \ref{fig:expsetup-occ} demonstrate the experimental setup of the pre-study.

\FloatBarrier

\section{Additional Material from Experiment 1}
\label{sec:appex1}

Figure \ref{fig:CRs_total} demonstrates the distribution of the answers for each CR in Experiment 1. The change in distribution is clearly visible between the CRs. Figure~\ref{fig:stacked-bars} provides an aggregated view of these results.
\begin{figure}[ht]
    \centering
    \resizebox{.5\linewidth}{!}{%
\begin{tikzpicture}[scale=1, font=\LARGE]

\definecolor{color0}{RGB}{0, 73, 131} 

\begin{groupplot}[group style={group size=2 by 2, horizontal sep=1.75cm, vertical sep = 1.75cm}]
\nextgroupplot[
tick align=outside,
tick pos=left,
title={CR-1, mean: 3.0168, std:1.2453},
x grid style={white!69.0196078431373!black},
xmin=-0.64, xmax=4.64,
xticklabels = {~,cmr,smr,eq,slr,clr},
xtick style={color=black},
y grid style={white!69.0196078431373!black},
ymin=0, ymax=2402.4,
ytick style={color=black}
]
\draw[draw=none,fill=color0] (axis cs:-0.4,0) rectangle (axis cs:0.4,1042);
\draw[draw=none,fill=color0] (axis cs:0.6,0) rectangle (axis cs:1.4,1529);
\draw[draw=none,fill=color0] (axis cs:1.6,0) rectangle (axis cs:2.4,2288);
\draw[draw=none,fill=color0] (axis cs:2.6,0) rectangle (axis cs:3.4,1543);
\draw[draw=none,fill=color0] (axis cs:3.6,0) rectangle (axis cs:4.4,1098);

\nextgroupplot[
tick align=outside,
tick pos=left,
title={CR-2, mean: 2.9929, std:1.223},
x grid style={white!69.0196078431373!black},
xmin=-0.64, xmax=4.64,
xticklabels = {~,cmr,smr,eq,slr,clr},
xtick style={color=black},
y grid style={white!69.0196078431373!black},
ymin=0, ymax=2530.5,
ytick style={color=black}
]
\draw[draw=none,fill=color0] (axis cs:-0.4,0) rectangle (axis cs:0.4,1025);
\draw[draw=none,fill=color0] (axis cs:0.6,0) rectangle (axis cs:1.4,1543);
\draw[draw=none,fill=color0] (axis cs:1.6,0) rectangle (axis cs:2.4,2410);
\draw[draw=none,fill=color0] (axis cs:2.6,0) rectangle (axis cs:3.4,1504);
\draw[draw=none,fill=color0] (axis cs:3.6,0) rectangle (axis cs:4.4,1018);

\nextgroupplot[
tick align=outside,
tick pos=left,
title={CR-4, mean: 3.0055, std:1.2539},
x grid style={white!69.0196078431373!black},
xmin=-0.64, xmax=4.64,
xticklabels = {~,cmr,smr,eq,slr,clr},
xtick style={color=black},
y grid style={white!69.0196078431373!black},
ymin=0, ymax=2283.75,
ytick style={color=black}
]
\draw[draw=none,fill=color0] (axis cs:-0.4,0) rectangle (axis cs:0.4,1077);
\draw[draw=none,fill=color0] (axis cs:0.6,0) rectangle (axis cs:1.4,1566);
\draw[draw=none,fill=color0] (axis cs:1.6,0) rectangle (axis cs:2.4,2175);
\draw[draw=none,fill=color0] (axis cs:2.6,0) rectangle (axis cs:3.4,1603);
\draw[draw=none,fill=color0] (axis cs:3.6,0) rectangle (axis cs:4.4,1079);

\nextgroupplot[
tick align=outside,
tick pos=left,
title={CR-8, mean: 2.9848, std:1.299},
x grid style={white!69.0196078431373!black},
xmin=-0.64, xmax=4.64,
xticklabels = {~,cmr,smr,eq,slr,clr},
xtick style={color=black},
y grid style={white!69.0196078431373!black},
ymin=0, ymax=1914.15,
ytick style={color=black}
]
\draw[draw=none,fill=color0] (axis cs:-0.4,0) rectangle (axis cs:0.4,1189);
\draw[draw=none,fill=color0] (axis cs:0.6,0) rectangle (axis cs:1.4,1681);
\draw[draw=none,fill=color0] (axis cs:1.6,0) rectangle (axis cs:2.4,1823);
\draw[draw=none,fill=color0] (axis cs:2.6,0) rectangle (axis cs:3.4,1669);
\draw[draw=none,fill=color0] (axis cs:3.6,0) rectangle (axis cs:4.4,1138);
\end{groupplot}

\end{tikzpicture}
    }
    \caption{The distribution of each stacked bar from Figure \ref{fig:stacked-bars}.}
    \label{fig:CRs_total}
\end{figure}
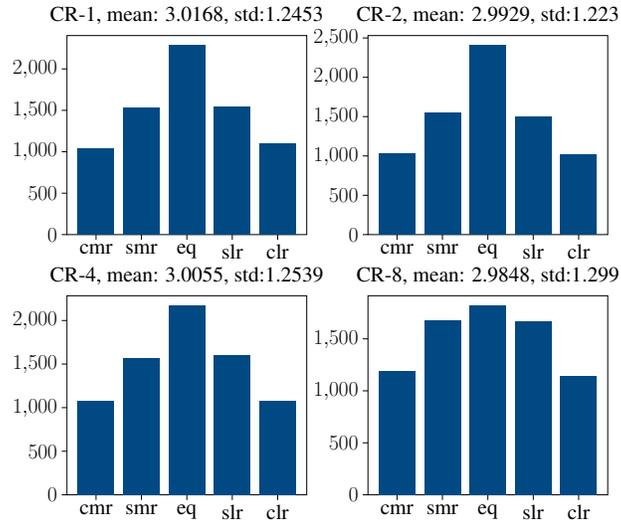
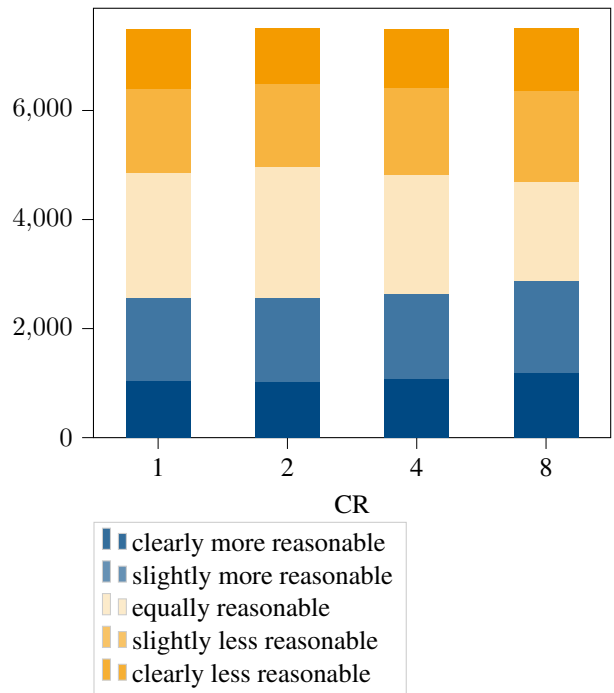
\begin{figure}[ht]
	\centering
	\resizebox{.5\linewidth}{!}{%
\begin{tikzpicture}

\definecolor{color0}{RGB}{0, 73, 131} 
\definecolor{color4}{RGB}{244, 155, 0} 
\colorlet{color1}{color0!75}
\colorlet{color2}{color4!25}
\colorlet{color3}{color4!75}

\begin{axis}[
legend cell align={left},
legend style={fill opacity=0.8, draw opacity=1, text opacity=1, draw=white!80!black,at={(0.0,-.40)},anchor=west},
legend cell align=left,
tick align=outside,
tick pos=left,
x grid style={white!69.0196078431373!black},
xmin=-0.5, xmax=3.5,
xtick style={color=black},
xtick={0,1,2,3},
xticklabels={1,2,4,8},
xlabel={CR},
y grid style={white!69.0196078431373!black},
ymin=0, ymax=7875,
ytick style={color=black}
]

\addlegendimage{ybar,ybar legend,draw=none,fill=color0}
\addlegendentry{clearly more reasonable}
\draw[draw=none,fill=color0] (axis cs:-0.25,0) rectangle (axis cs:0.25,1042);
\draw[draw=none,fill=color0] (axis cs:0.75,0) rectangle (axis cs:1.25,1025);
\draw[draw=none,fill=color0] (axis cs:1.75,0) rectangle (axis cs:2.25,1077);
\draw[draw=none,fill=color0] (axis cs:2.75,0) rectangle (axis cs:3.25,1189);

\addlegendimage{ybar,ybar legend,draw=none,fill=color1}
\addlegendentry{slightly more reasonable}
\draw[draw=none,fill=color1] (axis cs:-0.25,1042) rectangle (axis cs:0.25,2571);
\draw[draw=none,fill=color1] (axis cs:0.75,1025) rectangle (axis cs:1.25,2568);
\draw[draw=none,fill=color1] (axis cs:1.75,1077) rectangle (axis cs:2.25,2643);
\draw[draw=none,fill=color1] (axis cs:2.75,1189) rectangle (axis cs:3.25,2870);

\addlegendimage{ybar,ybar legend,draw=none,fill=color2}
\addlegendentry{equally reasonable}
\draw[draw=none,fill=color2] (axis cs:-0.25,2571) rectangle (axis cs:0.25,4859);
\draw[draw=none,fill=color2] (axis cs:0.75,2568) rectangle (axis cs:1.25,4978);
\draw[draw=none,fill=color2] (axis cs:1.75,2643) rectangle (axis cs:2.25,4818);
\draw[draw=none,fill=color2] (axis cs:2.75,2870) rectangle (axis cs:3.25,4693);

\addlegendimage{ybar,ybar legend,draw=none,fill=color3}
\addlegendentry{slightly less reasonable}
\draw[draw=none,fill=color3] (axis cs:-0.25,4859) rectangle (axis cs:0.25,6402);
\draw[draw=none,fill=color3] (axis cs:0.75,4978) rectangle (axis cs:1.25,6482);
\draw[draw=none,fill=color3] (axis cs:1.75,4818) rectangle (axis cs:2.25,6421);
\draw[draw=none,fill=color3] (axis cs:2.75,4693) rectangle (axis cs:3.25,6362);

\addlegendimage{ybar,ybar legend,draw=none,fill=color4}
\addlegendentry{clearly less reasonable}
\draw[draw=none,fill=color4] (axis cs:-0.25,6402) rectangle (axis cs:0.25,7500);
\draw[draw=none,fill=color4] (axis cs:0.75,6482) rectangle (axis cs:1.25,7500);
\draw[draw=none,fill=color4] (axis cs:1.75,6421) rectangle (axis cs:2.25,7500);
\draw[draw=none,fill=color4] (axis cs:2.75,6362) rectangle (axis cs:3.25,7500);
\end{axis}

\end{tikzpicture}
	}
	\caption{Aggregated results of Experiment 1 for all CRs.}
	\label{fig:stacked-bars}
\end{figure}
\FloatBarrier

\section{Additional Material from Experiment 2}
\label{sec:appex2}

\begin{figure}[H]
    \resizebox{.7\linewidth}{!}{%
	\input{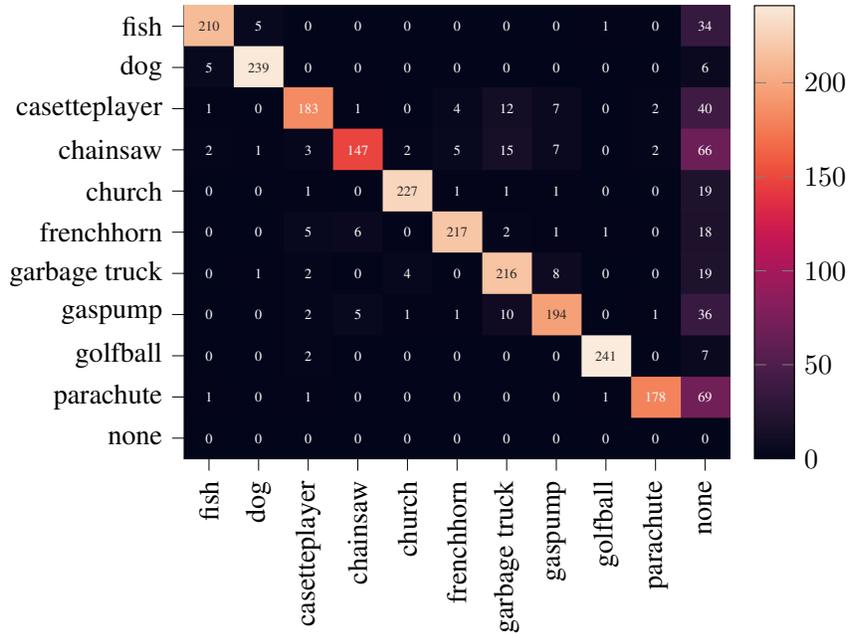}
    }
	\caption{Confusion Matrix CR 4.}
	\label{fig:exp2-conf-CR4}
\end{figure}

\begin{figure}[H]
\resizebox{.7\linewidth}{!}{%
  \input{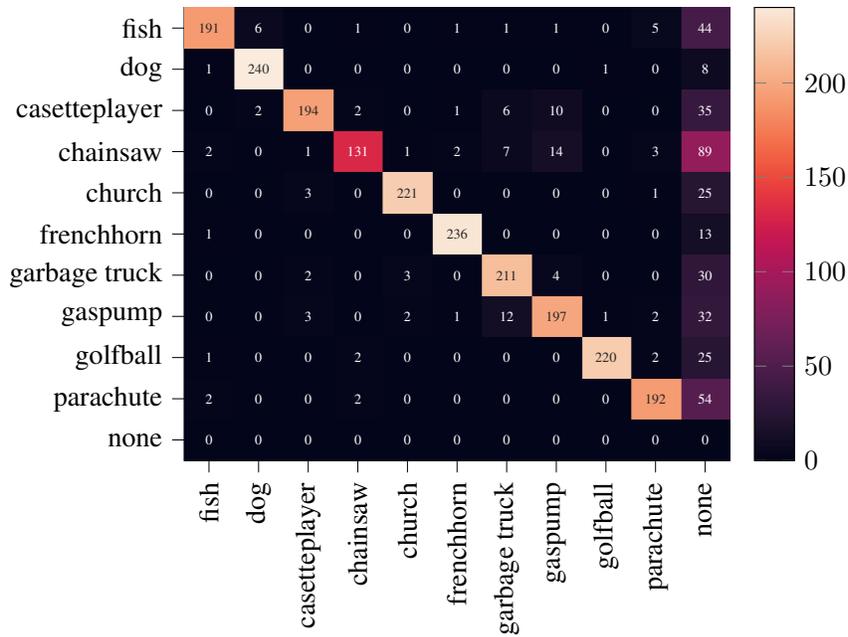}
  }
  \caption{Confusion Matrix CR 32.}
  \label{fig:confusion-matrix32}
  \end{figure}

Figures \ref{fig:exp2-conf-CR4} (CR 4), \ref{fig:exp2-conf-CR8} (CR 8), and \ref{fig:confusion-matrix32} (CR 32) complement the previous confusion matrices presented in section \ref{susec:experiment2}. Darker values indicate lower numbers, lighter values indicate higher numbers. The diagonals display correct classifications, while the right-most column shows the number of 'I don't know / None of the above'. 

Together with CR 8, CR 4 has the highest error-rate (5.36\%). The highest indecisiveness and error-rate is given for class 'chain saw' across all CR. CR 8 demonstrated the lowest accuracy in Experiment 2 (81.2\%). CR 32 achieved the highest indecisiveness (14.2\%) and second lowest accuracy in Experiment 2 (81.32\%). To summarize, table \ref{tab:idk_diffs} provides an overview of indecisiveness per class and CR.

\begin{figure}[H]
\resizebox{.7\linewidth}{!}{%
	\input{fig/conf-matrix-CR8}
    }
	\caption{Confusion Matrix CR 8.}
	\label{fig:exp2-conf-CR8}
\end{figure}

\begin{table}[tbh]
    \centering
    \begin{tabular}{lrrrrr}\toprule 
    {} & CR 1 & CR 2 & CR 4 & CR 8 & CR 32 \\ 
    \midrule
    fish            &   \textbf{6.8\%} &   7.2\% &  13.6\% &  18.4\% &   17.6\% \\ 
    dog             &   8.8\% &   4.0\% &   2.4\% &   \textbf{2.0}\% &    3.2\% \\ 
    cassette player  &  13.2\% &  \textbf{12.0\%} &  16.0\% &  17.6\% &   14.0\% \\ 
    chainsaw        &  22.4\% &  \textbf{22.0\%} &  26.4\% &  29.2\% &   35.6\% \\ 
    church          &  12.8\% &   \textbf{3.6\%} &   7.6\% &   9.6\% &   10.0\% \\ 
    french horn      &   4.0\% &   \textbf{3.2\%} &   7.2\% &   6.0\% &    5.2\% \\ 
    garbage truck    &   7.2\% &   \textbf{5.2\%} &   7.6\% &   9.6\% &   12.0\% \\
    gas pump         &  \textbf{11.6\%} &  14.8\% &  14.4\% &  16.0\% &   12.8\% \\ 
    golfball        &   2.4\% &   \textbf{1.6\%} &   2.8\% &   6.0\% &   10.0\% \\
    parachute       &   \textbf{8.8\%} &  16.0\% &  27.6\% &  20.0\% &   21.6\% \\ 
    \midrule 
    total           &  9.80\% &     \textbf{8.96\%} &    12.56\% &   13.44\% & 14.20\% \\
    \bottomrule
    \end{tabular}
    \caption{Indecisiveness of the respondents per class for all CRs.}
    \label{tab:idk_diffs}
\end{table}

\end{document}